\documentclass[conference]{IEEEtran}
\IEEEoverridecommandlockouts
\usepackage{cite}
\usepackage{amsmath,amssymb,amsfonts}
\usepackage{algpseudocode}
\usepackage{algorithm}
\usepackage{graphicx}
\usepackage{booktabs}
\usepackage{textcomp}
\usepackage{xcolor}
\usepackage{float}
\usepackage[hyphens]{url}
\usepackage{hyperref}  

\usepackage[caption = false]{subfig}
\def\BibTeX{{\rm B\kern-.05em{\sc i\kern-.025em b}\kern-.08em
    T\kern-.1667em\lower.7ex\hbox{E}\kern-.125emX}}
\begin{document}

\newcommand{\orcid}[1]{\href{https://orcid.org/#1}{#1}}

\title{
Faster Convergence for Transformer Fine-tuning with Line Search Methods
\thanks{We gratefully acknowledge the funding by the German Federal Ministry of Economic Affairs and Energy (01MK20007E). The work was
partially conducted within the KI-Starter project
“Robustness and Generalization in Training Deep Neural Networks” funded by the Ministry of Culture and Science Nordrhein-Westfalen, Germany}
}

\author{\IEEEauthorblockN{1\textsuperscript{st} Philip Kenneweg}
\IEEEauthorblockA{\textit{AG Machine Learning} \\
\textit{University Bielefeld}\\
Bielefeld, Germany \\
 \orcid{0000-0002-7097-173X}}
\and
\IEEEauthorblockN{2\textsuperscript{nd} Leonardo Galli}
\IEEEauthorblockA{\textit{MIP Chair} \\
\textit{RWTH Aachen University}\\
Aachen, Germany \\
\orcid{0000-0002-8045-7101}}
\and
\IEEEauthorblockN{3\textsuperscript{rd} Tristan Kenneweg}
\IEEEauthorblockA{\textit{AG Machine Learning} \\
\textit{University Bielefeld}\\
Bielefeld, Germany \\
 \orcid{0000-0001-8213-9396}}
\and
\IEEEauthorblockN{4\textsuperscript{th} Barbara Hammer}
\IEEEauthorblockA{\textit{AG Machine Learning} \\
\textit{University Bielefeld}\\
Bielefeld, Germany \\
 \orcid{0000-0002-0935-5591}}}

\maketitle

\begin{abstract}
Recent works have shown that line search methods greatly increase performance of traditional stochastic gradient descent methods on a variety of datasets and architectures \cite{armijo,vaswani20a}. In this work we succeed in extending line search methods to the novel and highly popular Transformer architecture and dataset domains in natural language processing. 
More specifically, we combine the Armijo line search with the Adam optimizer and extend it by subdividing the networks architecture into sensible units and perform the line search separately on these local units. 
Our optimization method outperforms the traditional Adam optimizer and achieves significant performance improvements for small data sets or small training budgets, while performing equal or better for other tested cases.
Our work is publicly available as a python package, which provides a hyperparameter-free pytorch optimizer that is compatible with arbitrary network architectures.
\end{abstract}


\section{\uppercase{Introduction}}
\label{sec:intro}

In modern machine learning, there is a variety of optimization algorithms. Figuring out which one is the best and matching the learning rate or learning rate schedule to the specific problem can require a lot of expert knowledge and computing power. In particular, the current state of the art is to treat the learning rate as a hyperparameter and train the network until the value that yields the best performance is found, i.e., hyperparameter tuning.
To simplify and significantly accelerate this process, a recent branch of the deep learning research \cite{mahsereci15a,bollapragada18a,paquette20a,armijo} has suggested to reintroduce line search methods for selecting the step size. These methods are widely adopted in the optimization community since their introduction in \cite{armijo66a}. In particular, they automatically find an adaptive learning rate by calculating the loss at different points along the (gradient) direction and selecting an appropriate step size. 
Using line search can save time and resources by eliminating the need for manual optimization of the learning rate, thus eliminating costly hyperparameter tuning.
\par As traditional line search requires multiple forward passes per gradient update, a more efficient approach is needed. In \cite{armijo}, a Stochastic Line Search (SLS) has been combined with a smart re-initialization of the step size to alleviate the need for multiple forward passes for every step.
This approach was shown to outperform a variety of other optimization methods, such as Stochastic Gradient Descent (SGD), SGD with Nesterov Momentum and many others on tasks such as matrix factorization as well as image classification for small networks and datasets.
\par It is however unclear if stochastic line search methods work well for larger networks and different architectures such as Transformers \cite{NIPS2017_3f5ee243} as well as in challenging domains such as Natural Language Processing (NLP). In this work we adapt and expand upon this active topic of research in the NLP domain.
\par In addition, given the deep structure of recent networks, it is natural to ask if it would be beneficial to train each layer at a different speed, i.e. learning rate, instead of using a single value for the whole network. For this reason, we propose to modify SLS \cite{armijo} and apply it in a layer-wise manner in order to specialize the step size for separate network components. This novel idea aims at localizing the line search in order to capture the loss variation due to the update of a single network component. 

\par The Adam \cite{adam} optimizer has been shown to outperform SGD in the training of Transformers \cite{anonymous2023heavytailed} by a large margin. In this work, we propose to use SLS \cite{vaswani20a} in combination with Adam to train Transformers or any other architecture. 
In particular, we will compare the following optimization schemas on the problem of fine-tuning Transformers:
\begin{itemize}
    \item Adam optimizer with warm starting and cosine decay (ADAM).
    \item Armijo line search combined with stochastic gradient descent (SGDSLS).
    \item Armijo line search combined with the Adam optimizer (ADAMSLS).
    \item Per-layer-ADAMSLS, as a novel optimization method (PLASLS).
\end{itemize}
We test all optimizers fine-tuning BERT\cite{bert} on the Glue\cite{wang-etal-2018-glue} dataset collection, which is widely used to evaluate common natural language processing skills.
We find that SGDSLS does not work well in this scenario.
In contrast, ADAMSLS and PLASLS perform better than ADAM, especially if they are trained on smaller data sets, as is often the case for retraining language models \cite{bert}.
To make our work easy to reproduce and easy to use, we implement all methods as pytorch optimizers. ADAMSLS can be used as an optimizer, that does not require to manually set the learning rate. For PLASLS the network parameters need to be manually split into components which are optimized separately.
\par In the next Section, we will review the related literature, in Section \ref{sec:methods} we will describe our method. In Section \ref{sec:experiments} we will describe our experimental approach and in Section \ref{sec:results} we will show and discuss our results. In Section \ref{sec:ablation} we will go into details on some choices for PLASLS. Finally Section \ref{sec:conclusion} contains our conclusion. 
%\par The loss function is denoted by $f(w) = \sum_{i=1}^{m} f_i(w)$, where $w\in\mathbb{R}^n$ are the parameters of the network, $m$ is the total amount of instances and $f_i$ is the loss corresponding to the instance $i$. With $|| \cdot ||$ we represent the Euclidean norm and $\nabla f$ is the gradient of $f$. Given $k$ the iteration counter, with $f_k$ and $\nabla f_k$ we denote the mini-batch function and its mini-batch gradient. 

\section{\uppercase{Related Work}} % focusing on evaluation of bias measures and fairness definitions in the literature
\label{sec:related_work}

%A large corpus of work has been published on gradient based optimization of neural networks. Stochastic gradient descend(SGD)

The optimization of deep neural networks has been a central topic of research in the field of machine learning. A variety of techniques and optimizers have been proposed, including but not limited to SGD \cite{robbins51a}, Adagrad\cite{adagrad}, RMSprop\cite{RMSprop} and Adam\cite{adam}. One common challenge in optimizing neural networks is the selection of appropriate step sizes, which can have a significant impact on the convergence rate and final performance of a model \cite{DBLP:journals/corr/abs-1805-08890}.

Line search methods are a popular approach for selecting step sizes in optimization algorithms. These methods involve iteratively adjusting the step size based on the curvature of the loss function, in order to ensure that the objective is decreasing at each iteration. However, traditional line search methods can be computationally expensive, particularly for large neural networks with many layers.

In this work we particularly build upon \cite{armijo}. In this paper, the authors show the theoretical proofs of convergence for the stochastic Armijo line search method in the case of strongly convex, convex and non convex functions. Furthermore, they provide solutions to some practical problems of line search methods. In particular, they no longer start the search from a fixed and possibly high initial step size, but instead from the last selected step size. To allow the step to also increase, they double it every $m/b$ steps, where $b$ is the minibatch size and $m$ is a constant that does normally not need to be tuned, see Eq. \ref{eq:tweak}. Additionally, they show empirical results on the image datasets MNIST\cite{mnist}, CIFAR10 \cite{cifar} and CIFAR100\cite{cifar} using the ResNet\cite{resnet} and DenseNet\cite{densenet} architecture, as well as on a variety of convex problems.

Important advantages of the Armijo line search compared to other optimization methods are:
\begin{itemize}
    \item no hyperparameter tuning of the learning rate
    \item faster convergence rates
    \item better generalization
\end{itemize}
However, their work does not show whether Armijo line search can outperform optimizers such as Adam on more complex tasks and architectures. 
To investigate this, we focus on natural language processing with Transformers, as this is a critical area of recent development and high complexity where other promising optimization methods have been unable to improve upon the baseline set by the Adam optimizer. Additionally, we investigate whether splitting the networks global step size into locally optimized step sizes is a viable option to further improve the performance of line search methods.

%To show this in particular, we focus on natural language processsing with transformers, as this is a critical area with a lot of recent developement and a high complexity, where other promising optimization methods have failed to improve the baseline set by the ADAM optimizer.

Recent work has shown that Transformers are very sensitive to the learning rate and learning rate schedule schedule they are trained with\cite{idealme, DBLP:journals/corr/abs-1908-03265}. To remedy this, various approaches like RADAM\cite{DBLP:journals/corr/abs-1908-03265} or warm starting have been proposed. We show that our approach is able to train these highly sensitive architectures well.

Other recent work includes optimizing the learning rate simultaneously with the network weights \cite{chandra2022gradient}. The authors showed that this yields good results on classical image datasets, but does not seem to speed up convergence.

Further related work includes \cite{DBLP:journals/corr/SinghDZGT15}, which focuses on layer specific learning rates, \cite{arous2022highdimensional} which studies the scaling limit of SGD in the high dimensional regime and \cite{anonymous2023heavytailed} which studies why Adam is so effective at training the Transformer architecture. 

Overall, the optimization of neural networks and transformer-based models remains an active area of research, and the development of efficient line search methods is an important contribution to this field.

\section{\uppercase{Methods}}
\label{sec:methods}
%We use ADAM and SGDSLS as baseline methods and present ADAMSLS and PLASLS as new techniques. This section contains short explanations of ADAM

The classical stochastic Armijo line search \cite{armijo} is designed to set a maximum step size for all network parameters $w_k$ at iteration $k$ (Eq. \ref{eq:armijo}).  In this section, together with the classical SGD direction, we will include the more complex Adam \cite{adam} direction. Moreover, we instead suggest to select a different step size for each different "component" of the network and evaluate the loss decrease w.r.t. it.

We define the following notation:
The loss function is denoted by $f(w)$. $|| \cdot ||$ denotes the Euclidean norm and $\nabla f$ denotes the gradient of $f$. Given the iteration counter $k$, $f_k$ and $\nabla f_k$ denote the mini-batch function and its mini-batch gradient. 

\subsection{Armijo Line Search} 
The Armijo line search criterion is defined in \cite{armijo} as:
\begin{equation}
    f_{k}(w_k + \eta_k d_k) \leq f_{k}(w_k) - c \cdot \eta_k ||\nabla f_{k}(w_k)||^2,
    \label{eq:armijo}
\end{equation}
where $d_k$ is the direction (e.g., $d_k=-\nabla f_{k}(w_k)$ in case of SGD), 
$c>0$ is a constant (commonly fixed to be $0.1$ \cite{armijo,anonymous2023heavytailed,vaswani20a}). Condition \ref{eq:armijo} is practically obtained by employing a backtracking procedure, i.e., starting with an initial step-size $\eta^0_k$ and iteratively decreasing it by a constant factor $\delta \in (0,1)$ until Eq. \ref{eq:armijo} is satisfied.
As described in Section \ref{sec:related_work}, the following tweak was introduced in \cite{armijo} to reduce the amount of backtracks, but avoid a monotonically decreasing step size
\begin{equation}\label{eq:tweak}
	\eta_k = \eta_{k-1} \cdot 2^{b/m}.
\end{equation}

\subsection{Including Adams Update Step in SLS (ADAMSLS)}

In case of SGD, the direction $d_k$ is the negative mini-batch stochastic gradient.
\begin{equation*}
     d_k = -\nabla f_{k}(w_k)
\end{equation*}
Adam's direction \cite{adam} can be defined as 
\begin{equation}
\begin{split}
   g_{k+1} &= \nabla f_{ik}(w_k) \\
   m_{k+1 } &= \beta_1 \cdot m_{k} + (1-\beta_1)\cdot g_{k+1} \\
   v_{k+1 } &= \beta_2 \cdot v_{k} + (1-\beta_2)\cdot g_{k+1}^2 \\
   \hat m_{k+1 } &=  m_{k} /(1-\beta^k_1) \\
   \hat v_{k+1 } &=  v_{k} /(1-\beta^k_2) \\
    d_k &= -\hat m_{k+1 } /(\sqrt{\hat v_{k+1 } }+\epsilon)
\end{split}
\label{eq:adamopt}
\end{equation}
Adam combines a momentum-based approach together with a step size correction built upon the gradients variance. In the training of Transformers, these modifications have been shown to be important enhancements over the simpler SGD\cite{arous2022highdimensional}. The weight update rule is generally defined as 
\begin{equation}
     w_{k+1} = w_{k} + \eta_k d_k.
\end{equation}
To apply the Armijo line search on the Adam optimizer we now use the direction $d_k$ defined in Eq. \ref{eq:adamopt}, but with momentum $\beta_1 = 0$. Additionally, the gradient norm term $||\nabla f_{k}(w_k)||^2$ is replaced by the scaled gradient norm $\frac{||\nabla f_{k}(w_k)||^2}{\sqrt{\hat v_{k+1 } }+\epsilon}$ resulting in Eq. \ref{eq:armijoadam}.
\begin{equation}
    f_{k}(w_k + \eta_k d_k) \leq f_{k}(w_k) - c \cdot \eta_k \frac{||\nabla f_{k}(w_k)||^2}{\sqrt{\hat v_{k+1 } }+\epsilon},
    \label{eq:armijoadam}
\end{equation}

\subsection{Layer Wise Line Search (PLASLS)}

The key idea of layer wise line search is to split up the network parameters into $L$ arbitrary sub-units, such as provided by network layers. In particular, we split $w_k$, $d_k$ and the gradient $\nabla f_{k}(w_k)$ into their components and rewrite $\eta_k$ as a vector:
\begin{equation*}
\begin{split}
w_k &= (w_k^{(1)}, \dots, w_k^{(l)},  \dots, w_k^{(L)})\\
\nabla f_{k}(w_k) &= \left({\nabla f_{k}(w_k)}^{(1)}, \dots, {\nabla f_{k}(w_k)}^{(l)}, \dots, {\nabla f_{k}(w_k)}^{(L)} \right)\\
d_k &= (d_k^{(1)}, \dots, d_k^{(l)},  \dots, d_k^{(L)})\\
\eta_k &= (\eta_k^{(1)}, \dots, \eta_k^{(l)},  \dots, \eta_k^{(L)}).
\end{split}
\end{equation*}
At this point, we adapt the Armijo line search to involve only elements of the sub-unit $(l)$. However, the loss $f$ is not partially separable \cite{galli20a}, i.e., it can only be computed on the whole set of parameters $w_k$. Therefore, we define gradient components $\bar{d}_{k,l}$ and rewrite  Eq. \ref{eq:armijo} as follows
\begin{equation}\label{eq:plsls}
\begin{split}
    f_{k}(w_k + \eta_k^{(l)} \bar{d}_{k,l}) &\leq f_{k}(w_k) - c \cdot \eta_k^{(l)} ||{\nabla f_{k}(w_k)}^{(l)}||^2,\\
    \bar{d}_{k,l} &:= (0, \dots, d_{k}^{(l)},  \dots, 0) 
\end{split}
    %\label{eq:armijo_per_section}
\end{equation}
This not only reduces the computational costs of the line search, but also localizes the method to capture the variation of the loss only w.r.t. the $l$-th sub-unit. At each iteration $k$, we only apply a line search procedure on a single sub-unit to yield $\eta_k^{(l)}$, while all the other step sizes remains unchanged. In other words, if $l$ is the sub-unit selected at iteration $k$ the update can be rewritten as 
\begin{equation} \label{eq:plasls}
\begin{split}
 w_{k+1} &= w_{k} + \eta_{k,l} \odot d_k,\\
\eta_{k,l} &:= (\eta_{k-1}^{(1)}, \dots, \eta_k^{(l)},  \dots, \eta_{k-1}^{(L)}),
\end{split}
\end{equation}
where $\odot$ is the Hadamard product, or component-wise product. Thus the resulting computational complexity is very similar to the original implementation, while allowing different step sizes for different network components.

%By reducing the the weight parameters which are considered in the Armijo line search criterion in Equation \ref{eq:plsls}, we keep all the convergence guaranties that the Armijo rule provides.

\subsubsection*{Splitting}
In the case of the Transformer\cite{NIPS2017_3f5ee243} architecture or more specifically the BERT\cite{bert} architecture which is encoder only, we split the network into 10 different components.
BERT has 12 encoder layers we divide the different layers into 8 different parts ($n_k^{(2)}$ affects layer 1, $n_k^{(3)}$ affects layers 2,3, $n_k^{(4)}$ affect layer 4, $n_k^{(5)}$ affect layers 5,6 ...).
$n_k^{(1)}$ affects the embedding weights of the network. $n_k^{(10)}$ affects the single dense layer which is appended for a specific task. This is the optimal split found as shown by the experiments in Section \ref{sec:ablation}.
% we divide the different layers into 8 different parts ($n_k^{(l)}$ affects the layer $l\!-\!1, \; \forall l=2, \dots, 9$). The first step size component 
% $(n_k^{(1)})$

%\subsubsection*{Efficiency}
%In practice it seems unnecessary to update the step size of each subnetwork $S^{(l)}$ each step. We choose to update a single subnetwork $S^{(l)}$ each step by cycling through the possibilities. With $l = l+1$ and if $l>L , l =0$ where $L$ is the number of subnetworks. %In Section \ref{sec:ablation} we investigate different update rules, but find no improvements upon this simple scheme.

\subsubsection*{First Step}
Since the step size for each sub-unit is not changed during each iteration, a large initial step size $\eta_0$ could lead to divergence problems.
To make our method completely automatic, we first perform a single step of SLS on the whole network to select a common first initial step size $\eta_0$. In particular, we choose a big initial step size $\eta_0=0.1$ and let the first line search find a good replacement for it. In fact, the step size yielded by the line search is a upper bound for the inverse of the Lipschitz constant of $\nabla f(w)$ (see Lemma 1 in \cite{armijo}) and this provides us with a good approximation of the complexity of the problem at hand (at least in the starting point). From this value, we then start subdividing the step sizes per sub-unit.
%The step size $n_k^{(l)}$ for each layer is only updated infrequently. But $n_k^{(l)}$ is initialized with a very high learning rate, this is problematic as during the first few steps, this can lead to divergence problems. To combat this we perform a global update step without the layer wise grouping first to obtain a reasonable step size and only then start with the described method.

\subsubsection*{Step Size Merging}
\begin{figure}
\includegraphics[width = 0.45\textwidth]{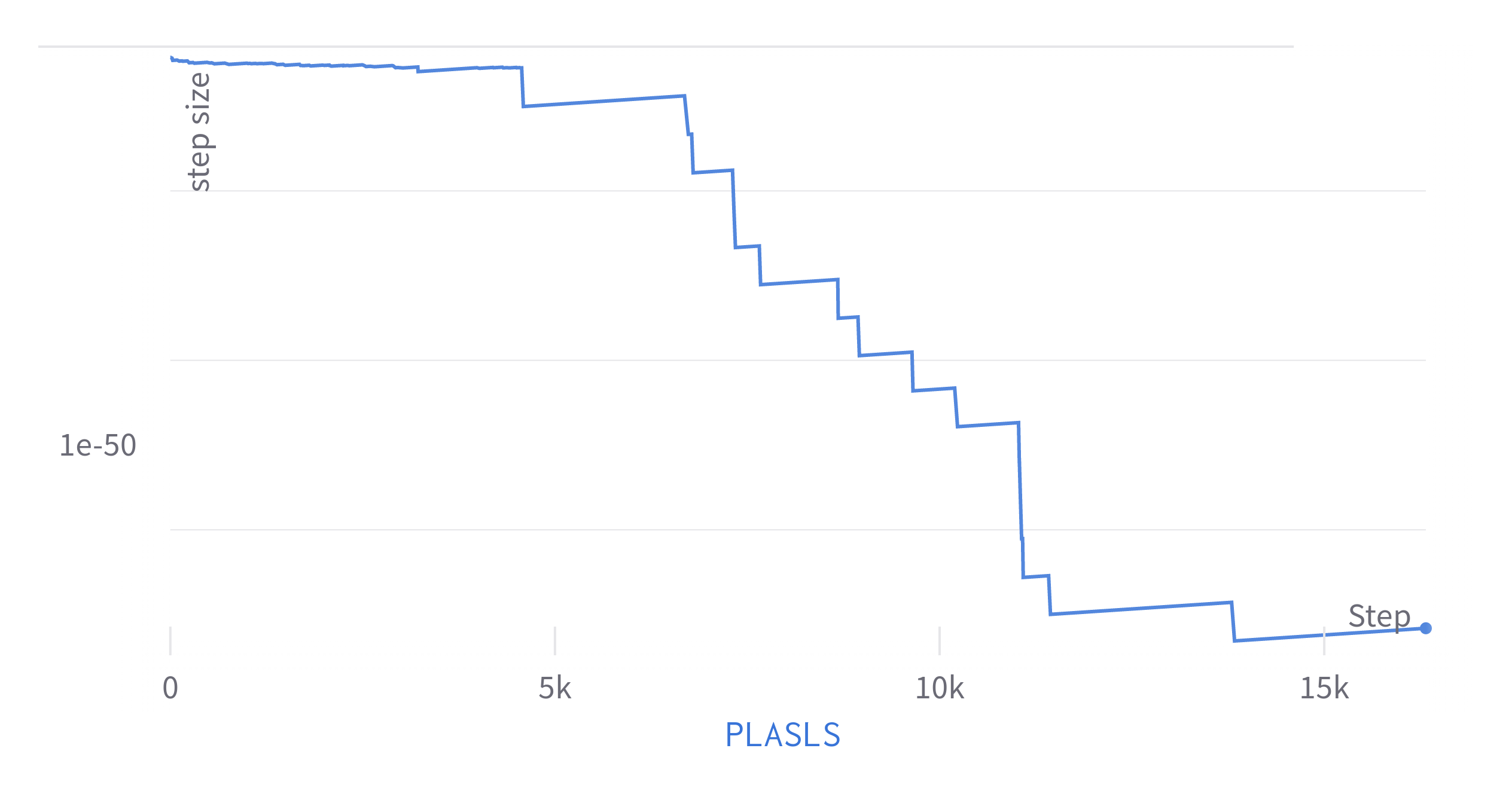}
\caption{Exemplary problematic step size of component 9 of the network during a single training run on the QNLI dataset. The step size starts out in the order of $10^{-4}$ but is lowered to values below $10^{-50}$. }
\label{fig:smallstep}
\end{figure}

If we look closely at the resulting step sizes yielded by our methods, we observe that they might converge to very low values (e.g., Figure \ref{fig:smallstep}). This also happens without the layer wise splitting, but less frequently. As a result, the affected part of the network is no longer able to learn. This can be desirable in some cases, but seems problematic overall.

%If we look closely at the resulting step sizes for our layer wise line search, see Figure \ref{fig:smallstep}, we observe that some parts of it are converging to very low values, something that also happens without the layer wise splitting but less frequently. As a result, the affected part of the network is no longer able to learn. This can be desirable in some cases, but seems problematic overall. %In Section \ref{sec:ablation} we show that this effect decreases with increasing batch size.

%We suggest, that this is due to the noisiness of the training data combined with the layer splitting increasing the impact of this noise. %This noisiness coupled with only a small part of the network trying to satisfy the modified armijo criterion, does seem to result in very low learning rates for part of the network. %, even though these are sometimes desirable this 
%. Consequently, the armijo criterion leads to lowering the learning rate too much at some point. After this it is not able to quickly recover as the learning rate is only doubled every 300 steps. 
%For further research, it would be interesting to study how to make the armijo criterion more robust towards these outliers.
%We suggest that larger batch sizes during training would help alleviate these problems, see Section \ref{sec:ablation}, but due to hardware limitations we are only able to show that this problem becomes worse for smaller batch sizes.

%\subsection{merging network parts}
In the specific case of layer-wise line search optimizer, a first solution for this problem would be to merge the network parts based on their current step size. 
%This was tested due to some network parts step size being very small after longer training runs, see Fig. \ref{fig:smallstep}. 
%As a possible solution to this problem we designed the following algorithm:

\begin{algorithm}
\caption{algorithm for merging network components}\label{alg:cap}
\begin{algorithmic}
\State step sizes $n_{k}^{(l)}$, threshold $\lambda$
\State find the smallest step size $n_{k}^{(s)} \in n_{k}^{(l)}$
\If{$n_{k}^{(s)} \leq \lambda$}
    \State find the second smallest step size $n_{k}^{(s2)} \in n_{k}^{(l)}$
    \State merge network components $s$ and $s2$
    \State $n_{k}^{(new)} = (n_{k}^{(s)} + n_{k}^{(s2)}) / 2$
\EndIf
\end{algorithmic}
\end{algorithm}
In Algorithm \ref{alg:cap}, we detect if a network components step size $n_{k}^{(l)}$ is below a certain threshold $\lambda$. If this is the case, this network component is merged with the network component with the second lowest step size. The new step size for this combined network component is the average of both step sizes.

\begin{figure}
\includegraphics[width = 0.45\textwidth]{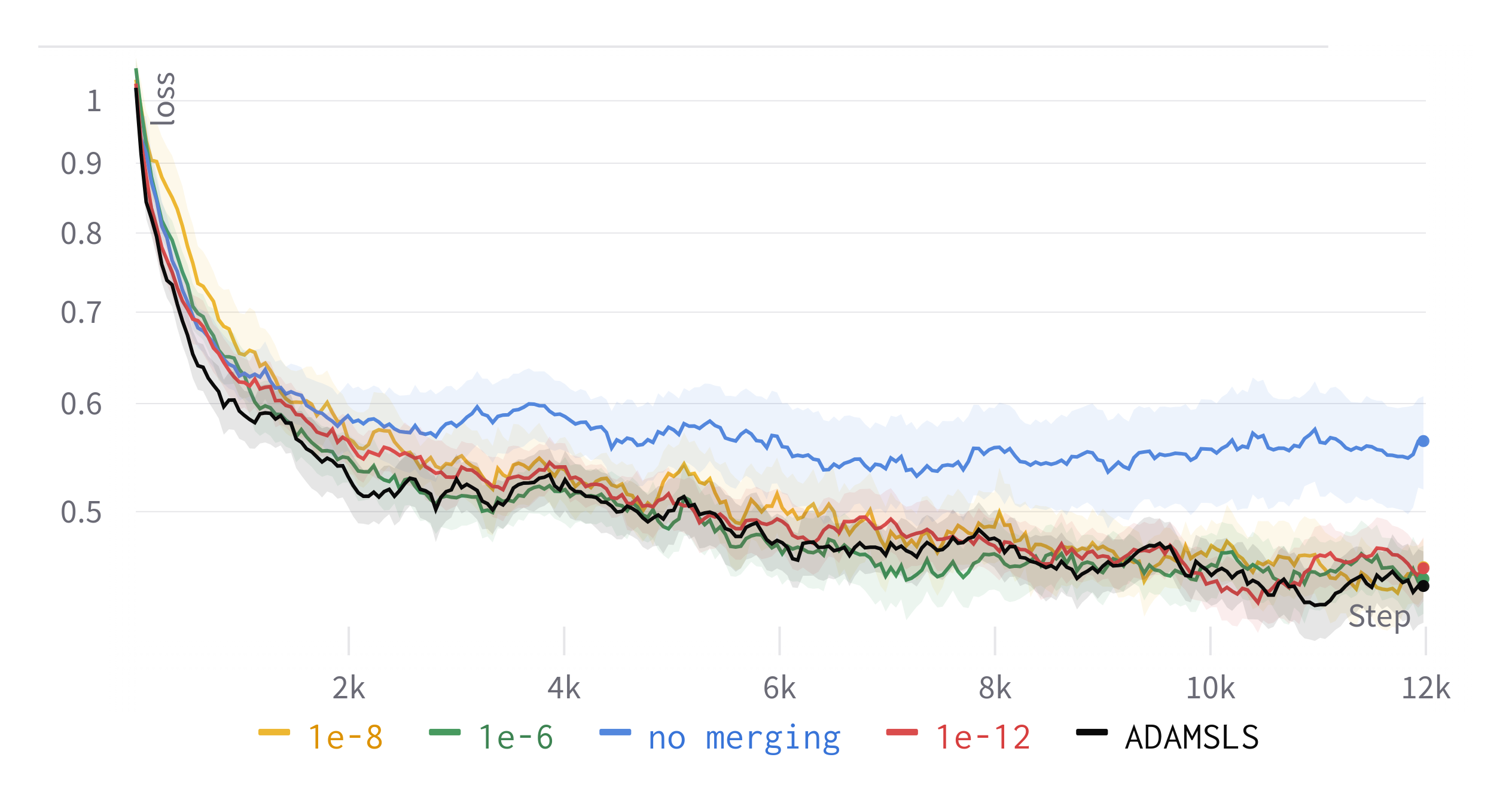}
\caption{Different merging thresholds on one epoch of the MNLI dataset. Standard error is indicated around each line.}
\label{fig:merge}
\end{figure}

In Figure \ref{fig:merge}, the results of different merging thresholds $\lambda$ are visualized, they seem to indicate that no merging performs worse than even the smaller merging thresholds of for example $\lambda = 10^{-12}$. Consequently, all experiments in Section \ref{sec:experiments} for PLASLS are conducted using the merging threshold $\lambda = 10^{-12}$.
With this threshold only very little automatic merging occurs. On the smaller datasets (all $small$ datasets, MRPC) automatic merging does not occur, while on the larger datasets (SST2, QNLI, MNLI) on average about 3 of the 10 components are merged during a full 5 epoch training run. We strongly suspect that numerical cancellation errors are the cause of very low step sizes, we therefore plan to investigate this topic in more details in the future.

%\subsection{Usability}
%To make our work easy to reproduce and easy to use, we implement all methods as pytorch optimizers. ADAMSLS can be used as a optimizer, albeit without the need to manually set the learning rate. For PLASLS the network parameters need to be split into parts which are optimized seperately. The approach of pytorch, to use different optimizers is very computationally inefficient in the PLASLS case. For this reason we initialize the PLASLS optimizer differently. Instead of a list of model parameters $w$, we use a nested list of model parameters $w^{(l)}$. It is thereby easy and intuitive to split the model parameters in any way. The splits presented in this paper are fitted to the transformer architecture, but the presented optimizer is able to work with any split.

\section{\uppercase{Experimental approach}}
\label{sec:experiments}
In this section, we detail our experimental design to investigate the performance of the proposed optimizers.
We utilize the Huggingface library \cite{https://doi.org/10.48550/arxiv.1910.03771} for implementation and the pre-trained Bert model 'bert-base-uncased' for all experiments.

\subsection{Candidates} 
As a baseline comparison we evaluate the Adam optimizer with a peak learning rate of 2e-5 and a cosine decay with warm starting for 10\% of the total training time, henceforth referenced to as $ADAM$. These values are taken from the original paper \cite{bert} and present good values for a variety of classification tasks, including the Glue\cite{wang-etal-2018-glue} tasks upon which we are evaluating.

As another baseline, we use Armijo line search as described in \cite{armijo} with stochastic gradient descent, we further refer to this approach as $SGDSLS$.

Next, we use the Armijo line search described in \cite{armijo} and combine it with the Adam optimizer\cite{adam} as described in Section \ref{sec:methods} to obtain the approach further referenced to as $ADAMSLS$.

The last option we are testing is the layer wise line search as described in Section \ref{sec:methods} combined with the Adam optimizer further referenced to as $PLASLS$. 

\subsection{Implementation Details} 
The metaparameter choices for all experiments are as follows:
\begin{itemize}
    \item All models are trained for 5 epochs.
    \item The pooling operation used in all experiments is [CLS]. 
    \item Batch size used for training is 32.
    \item The Adam optimizer with betas (0.9,0.999) and epsilon $10^{-8}$ is used.
    \item The maximum sequence length is set to 256 tokens.
    \item All models are trained 5 times and their mean metrics and standard error are reported.
\end{itemize}

\subsection{Datasets}

%\subsubsection{Glue}
We consider a common scenario in natural language processing, where a large pre-trained language model is fine-tuned on a small dataset.
The Glue dataset by Wang et al. \cite{wang-etal-2018-glue} is a collection of various popular classification tasks in NLP, and it is widely used to evaluate common natural language processing capabilites. All datasets used are the version provided by tensorflow-datasets 4.0.1.

%is done to research the question of different data set sizes requiring different classification head architectures for fine tuning.
More specifically, we use the datasets Stanford Sentiment Treebank \emph{SST2},
 Microsoft Research Paraphrase Corpus \emph{MRPC}, Stanford Question Answering Dataset \emph{QNLI},
 %the Corpus of Linguistic Acceptability \emph{COLA},
 and Multi-Genre Natural Language Inference Corpus 
  \emph{MNLI}.

   We evaluate all approaches on the $small$ and full size datasets. The $small$ datasets are the same as previously described, except that the size of the training dataset has been reduced to 500 randomly drawn samples. This scaling enables us to judge the capability of the optimizer in different dataset size regimes, as especially in real world scenarios small dataset sizes are a common occurrence.

\begin{figure}
\subfloat[MNLI small]{\includegraphics[width = 0.25\textwidth]{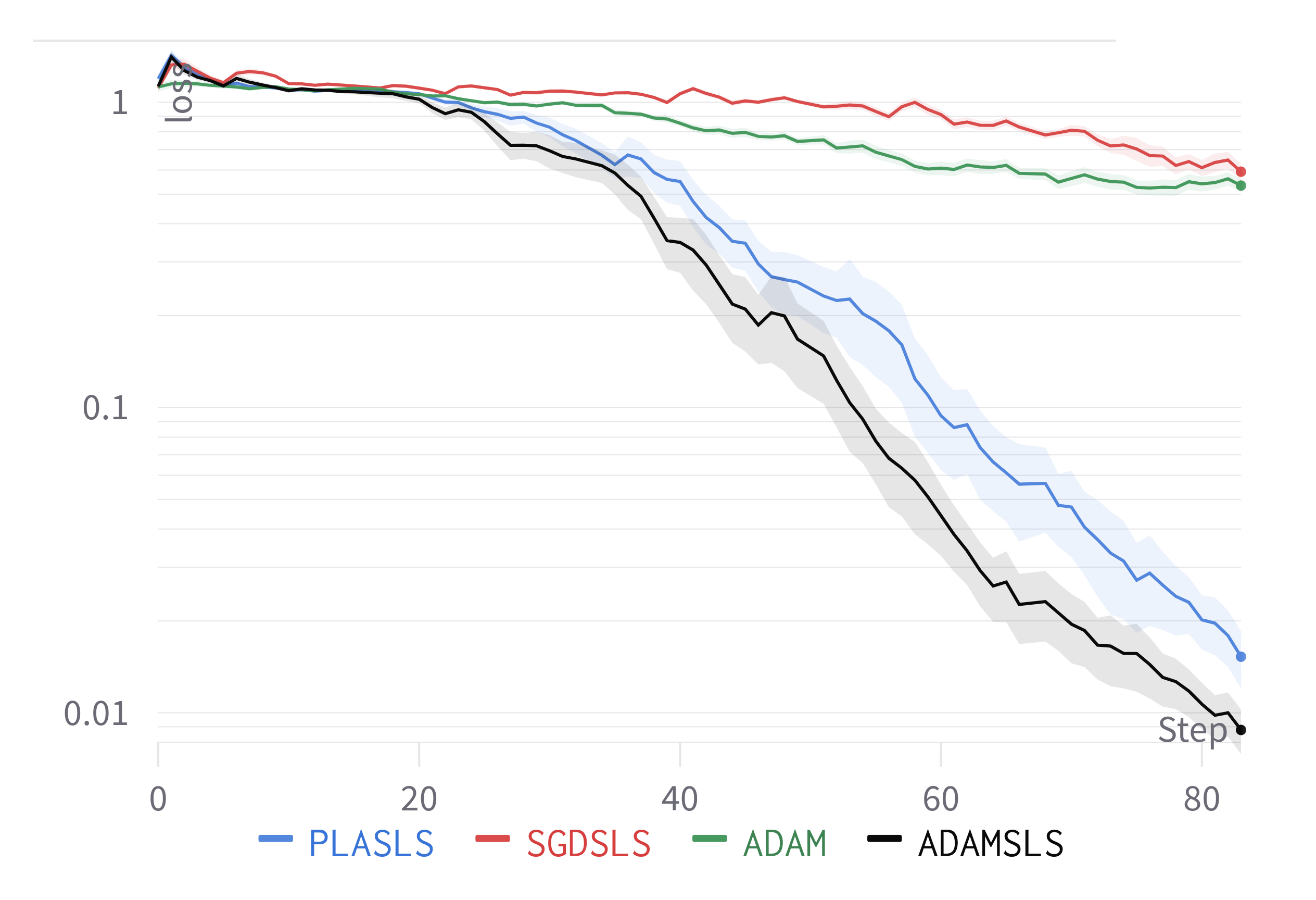}} 
\subfloat[MRPC small]{\includegraphics[width = 0.25\textwidth]{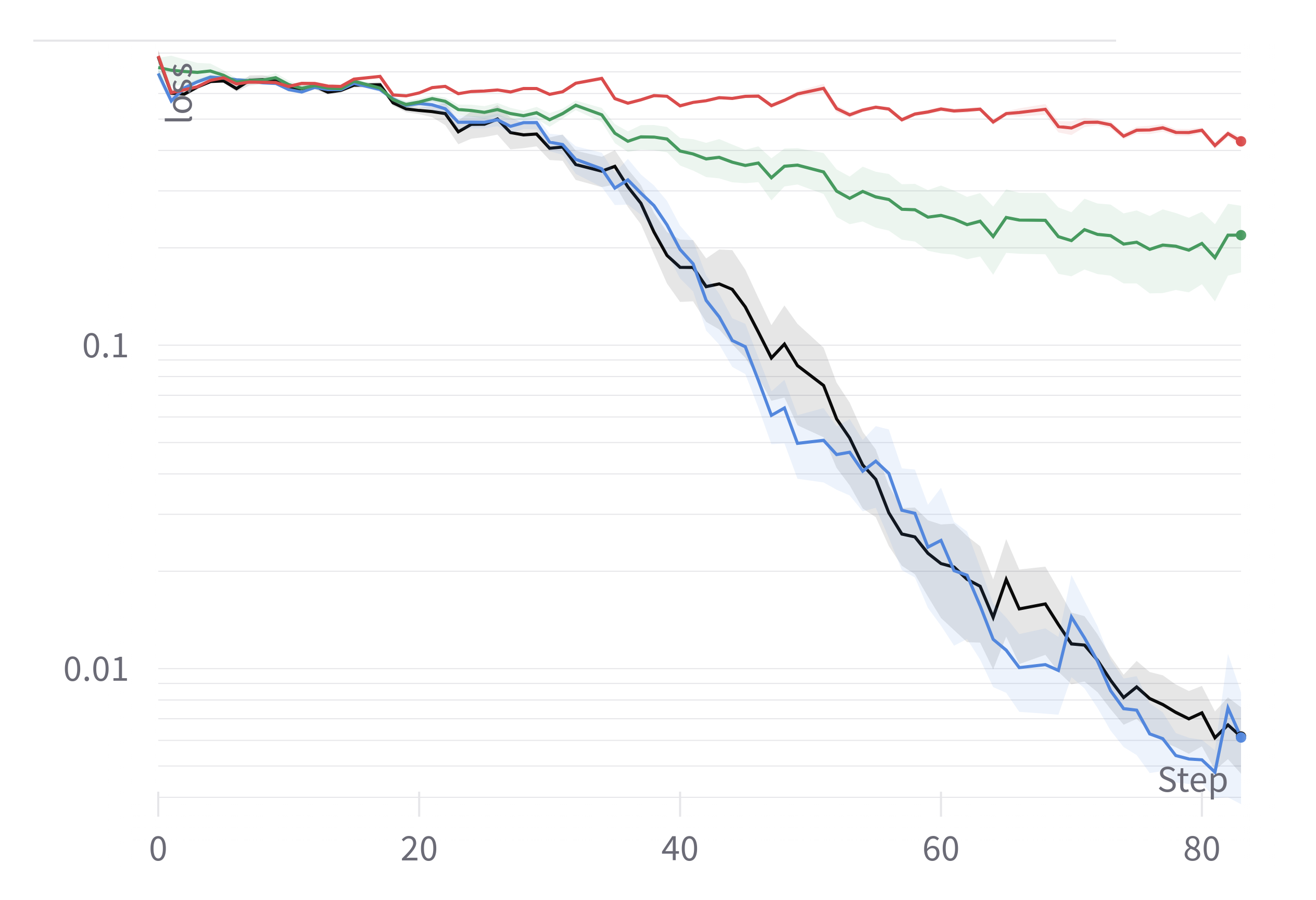}}\\
\subfloat[QNLI small]{\includegraphics[width = 0.25\textwidth]{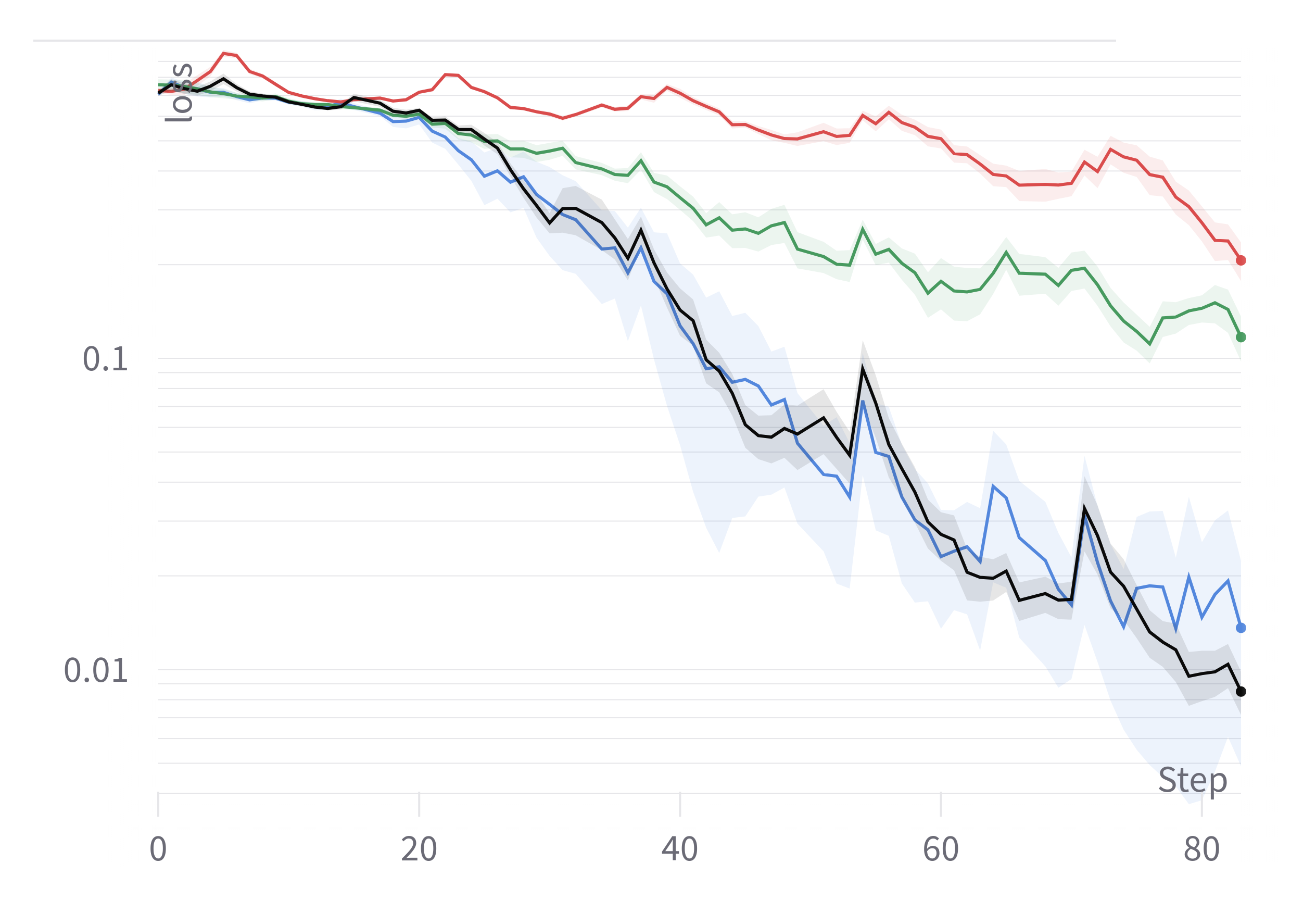}}
\subfloat[SST2 small]{\includegraphics[width = 0.25\textwidth]{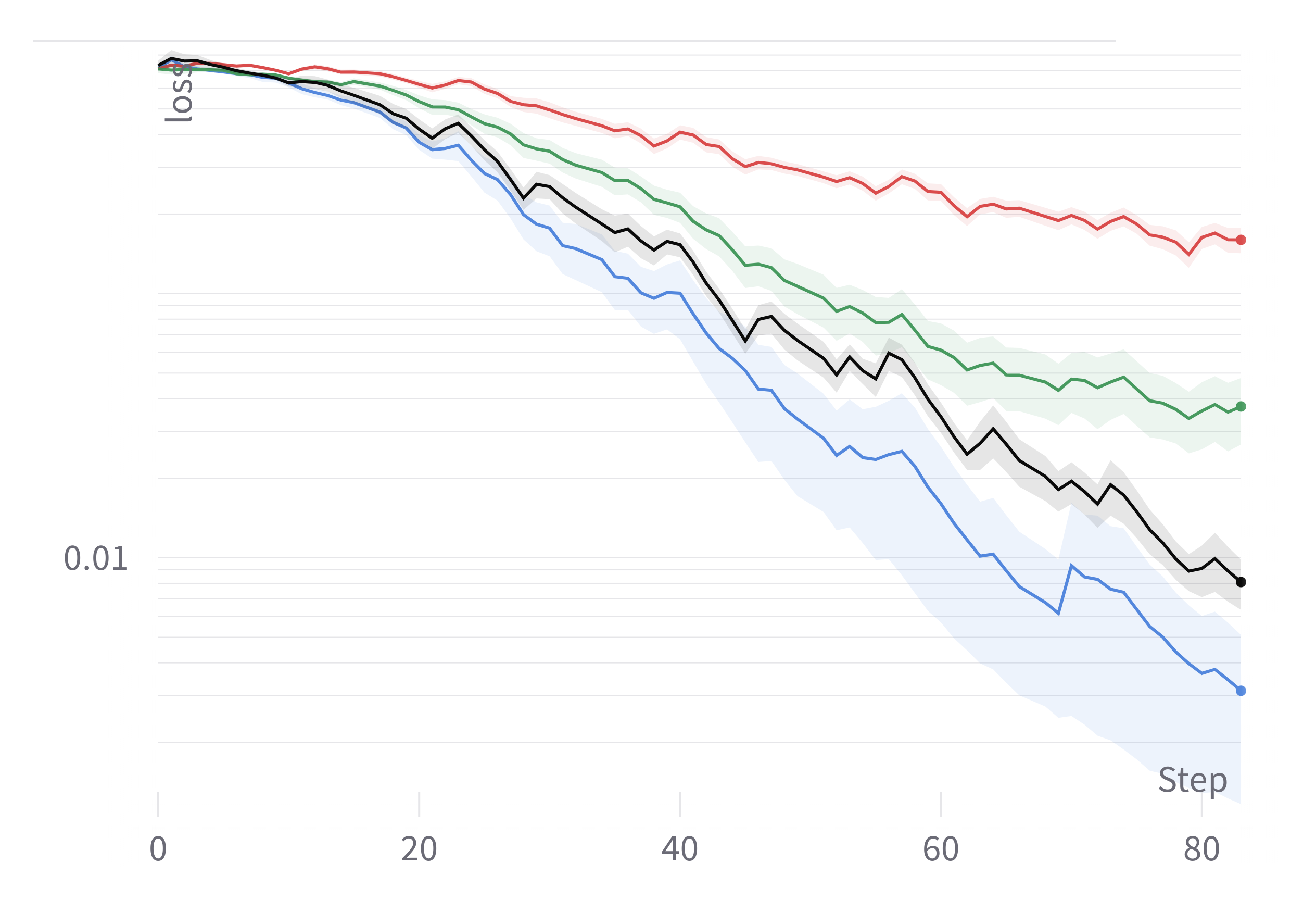}} 
\caption{The loss curves of the experiments on the small dataset with standard error indicated around each line. In all experiments SGDSLS performs the worst, followed by ADAM. PLASLS and ADAMSLS do not perform very different. In the SST2 and QNLI experiment PLASLS performs best, while in the MNLI experiment ADAMSLS performs best. In the MRPC experiments both perform about the same.}
\label{fig:smallexploss}
\end{figure}

\section{\uppercase{Experimental results}}
\label{sec:results}

\begin{figure}
\subfloat[MNLI small]{\includegraphics[width = 0.25\textwidth]{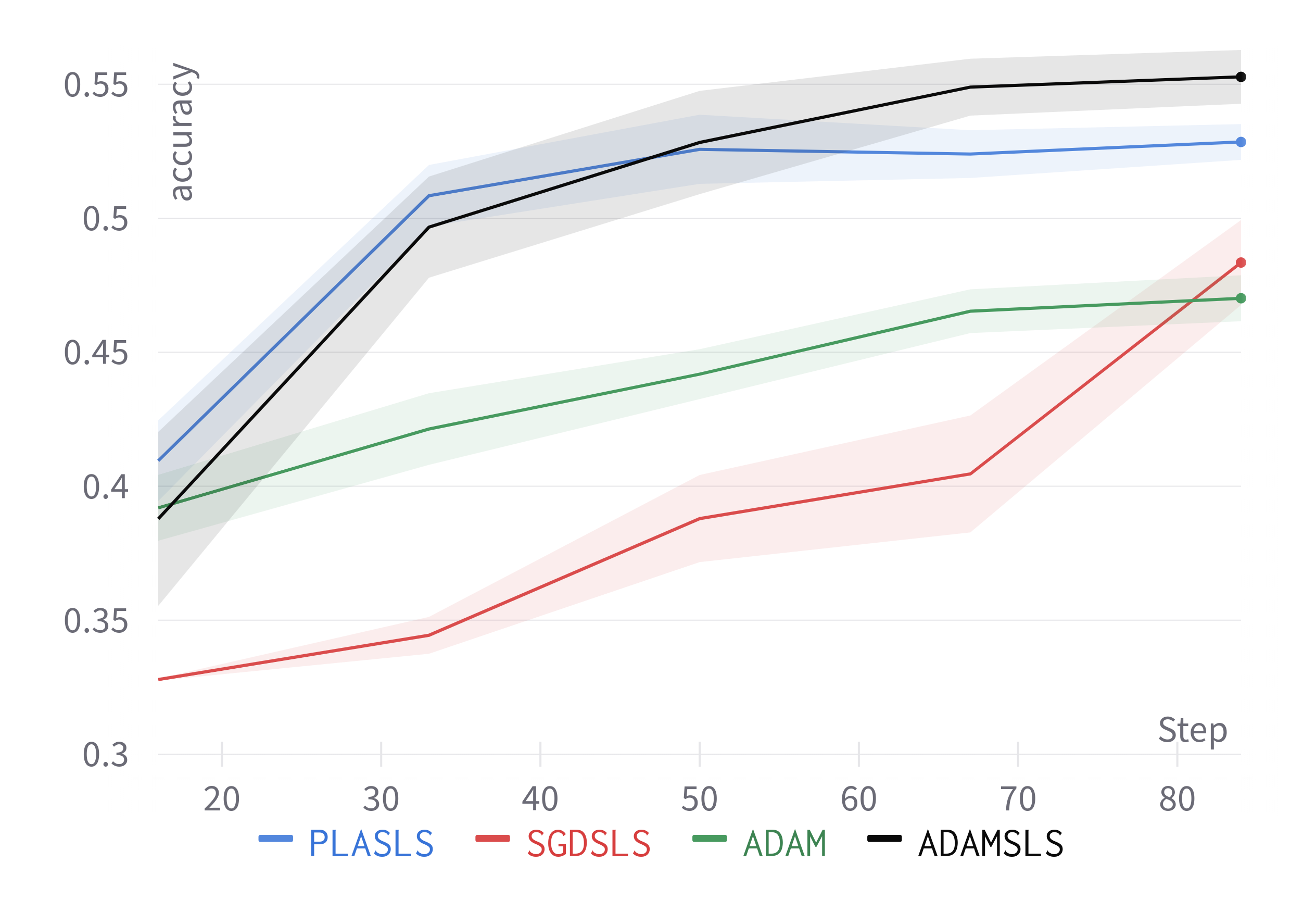}} 
\subfloat[MRPC small]{\includegraphics[width = 0.25\textwidth]{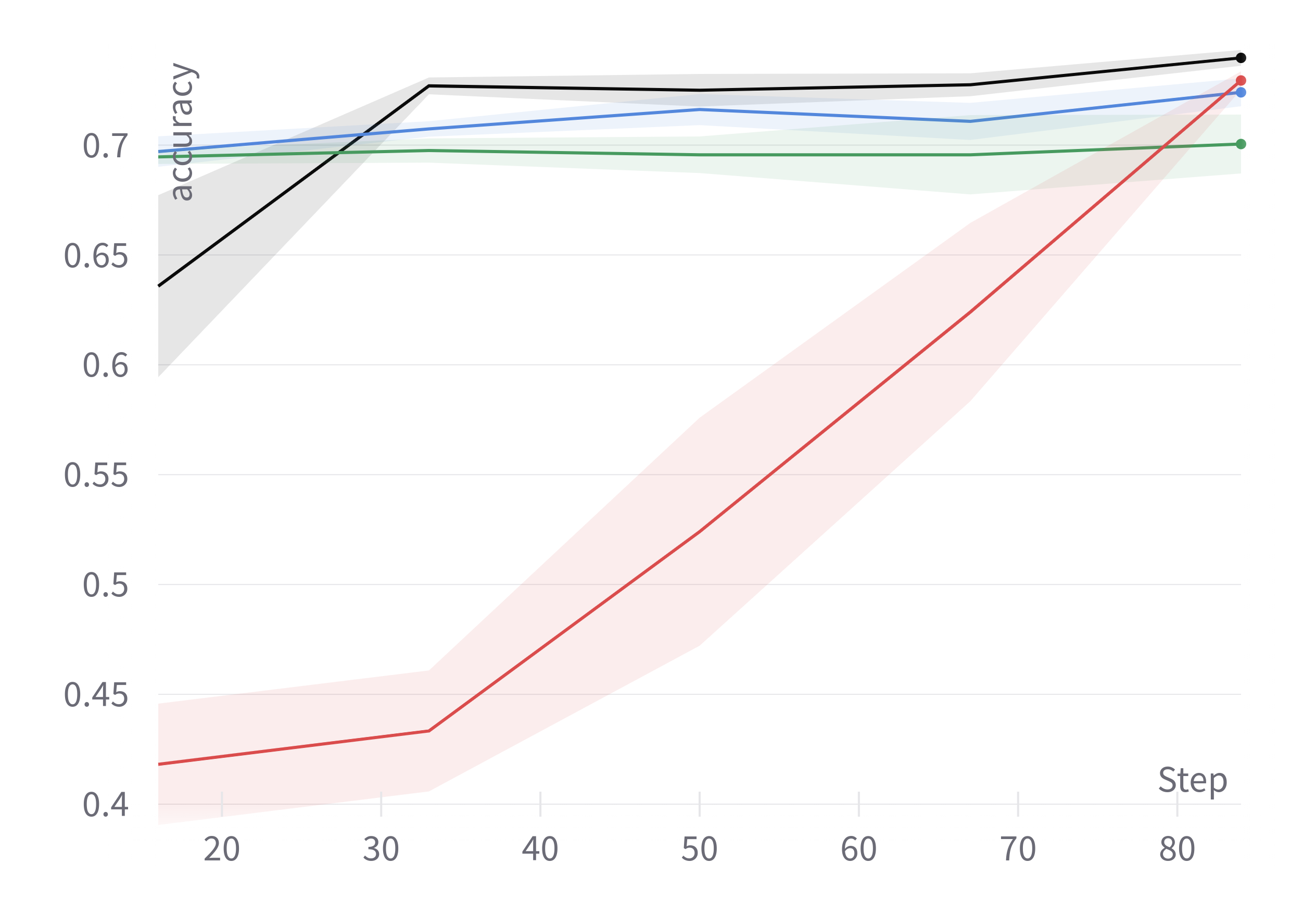}}\\
\subfloat[QNLI small]{\includegraphics[width = 0.25\textwidth]{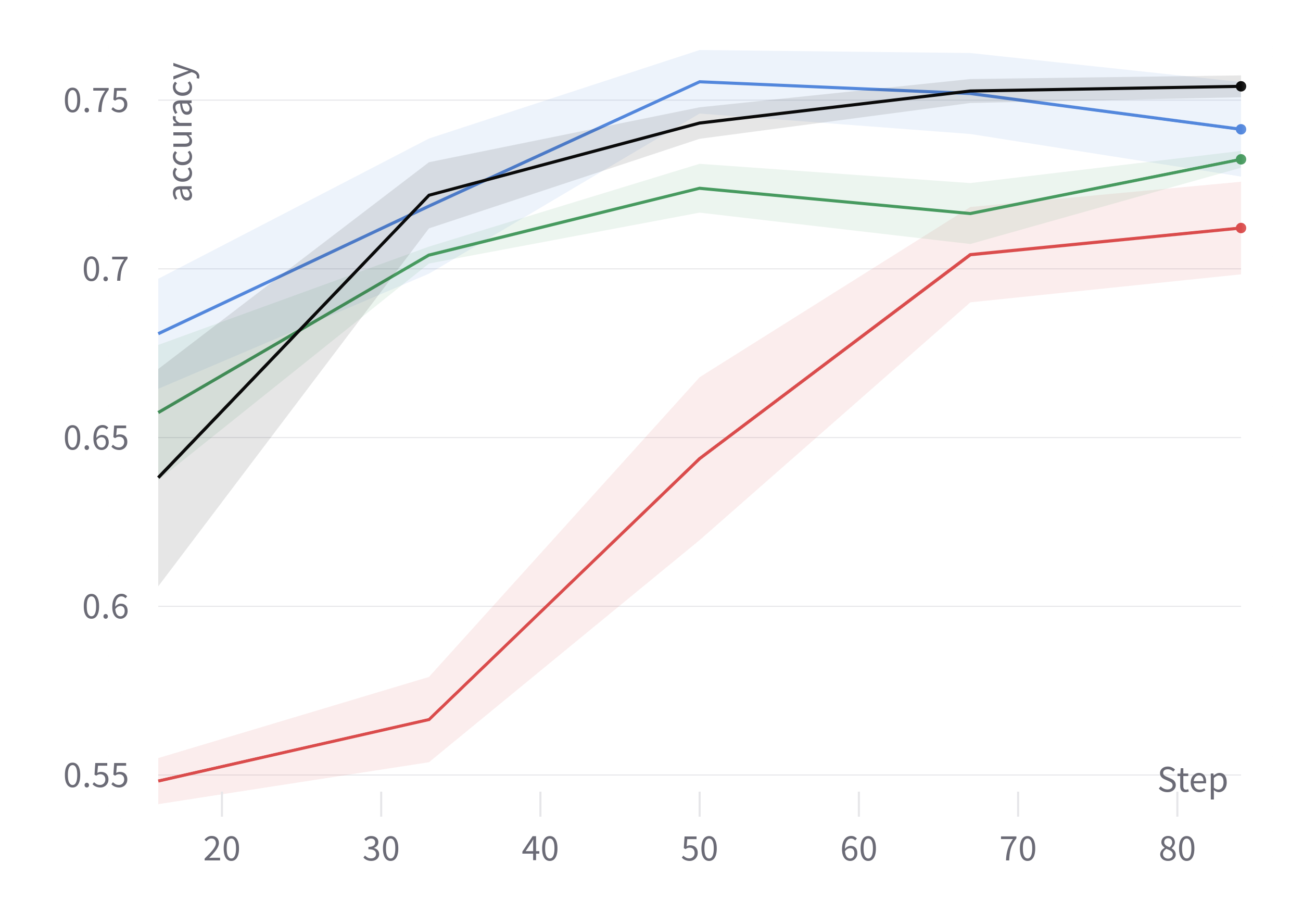}}
\subfloat[SST2 small]{\includegraphics[width = 0.25\textwidth]{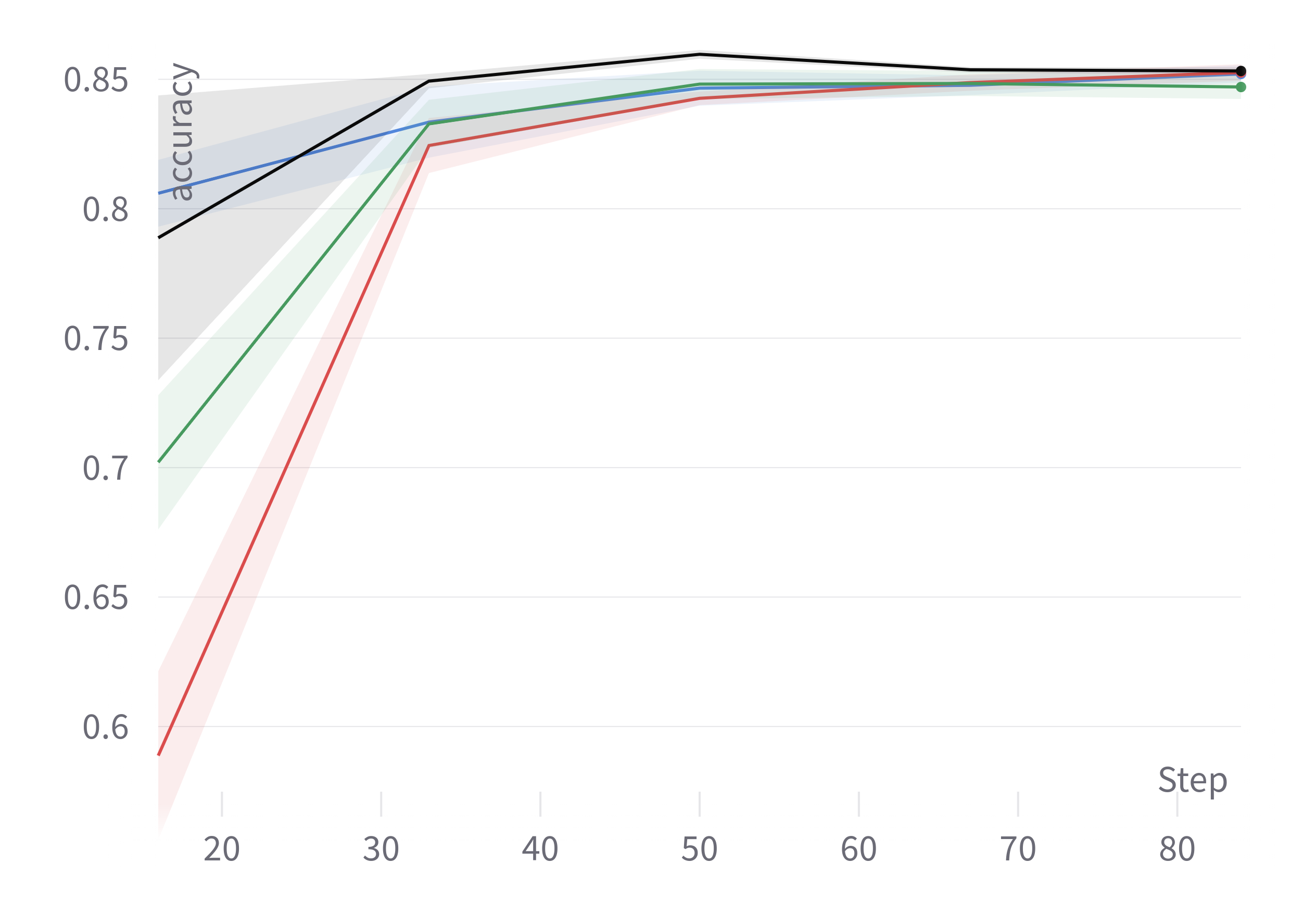}} 
\caption{The accuracy curves of the experiments on the small dataset with standard error indicated around each line, starting after the first epoch. In all experiments SGDSLS performs the worst, followed by ADAM. PLASLS and ADAMSLS do not perform very different. In the MNLI and MRPC experiment ADAMSLS performs best. In the SST2 and QNLI experiments ADAMSLS and PLASLS perform about the same.}
\label{fig:smallexpacc}
\end{figure}

\begin{figure}
\subfloat[MNLI ]{\includegraphics[width = 0.25\textwidth]{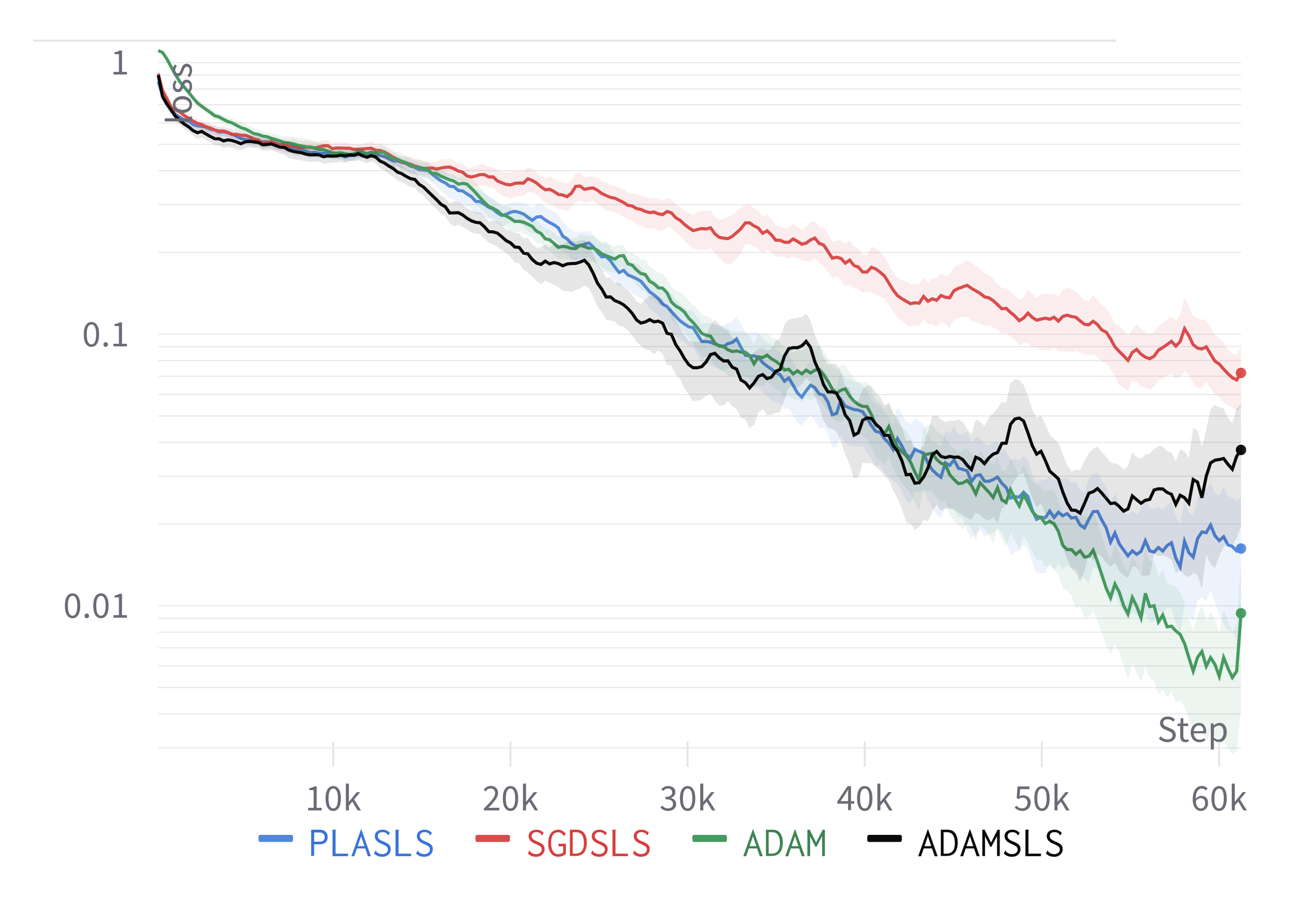}} 
\subfloat[MRPC ]{\includegraphics[width = 0.25\textwidth]{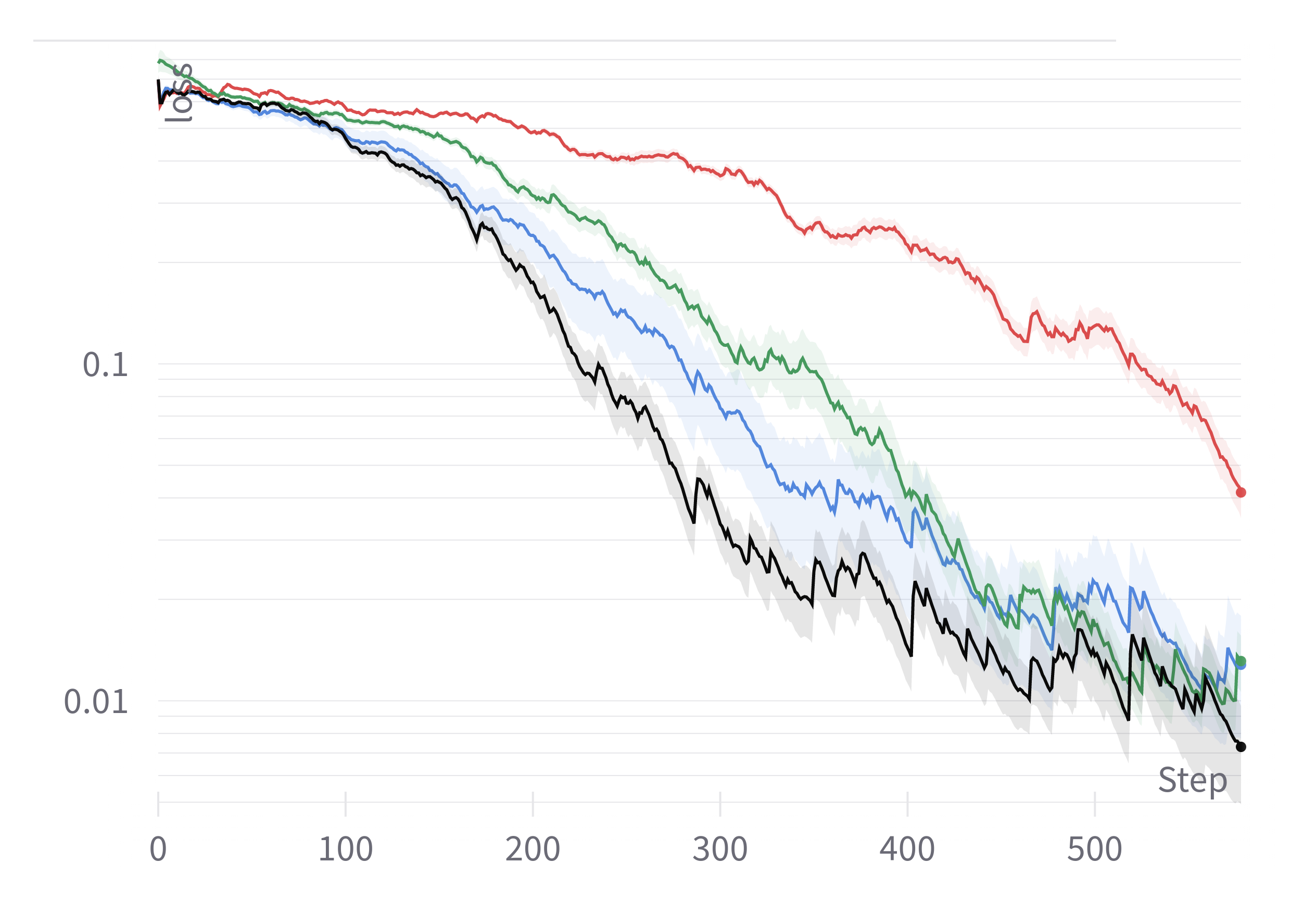}}\\
\subfloat[QNLI ]{\includegraphics[width = 0.25\textwidth]{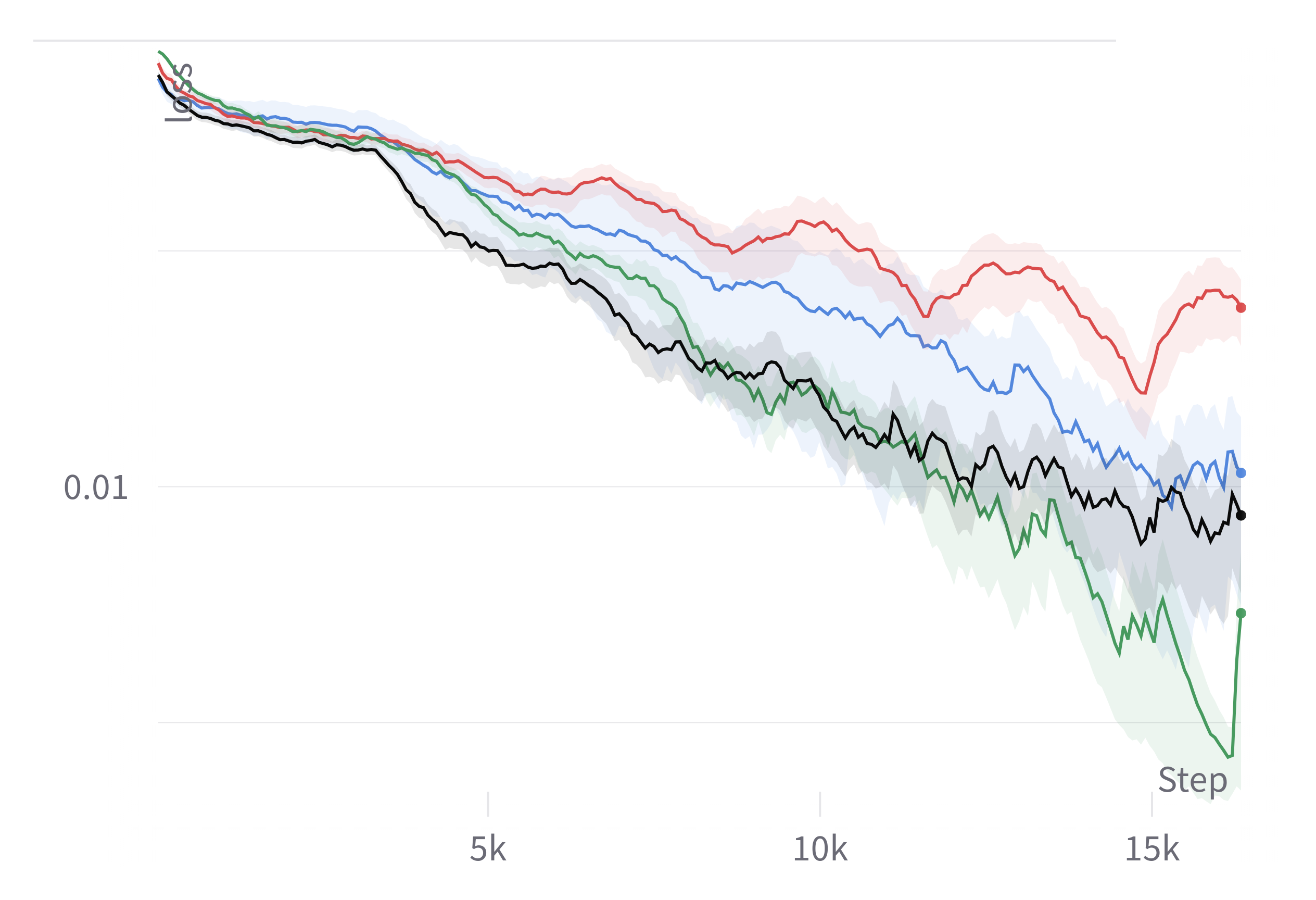}}
\subfloat[SST2 ]{\includegraphics[width = 0.25\textwidth]{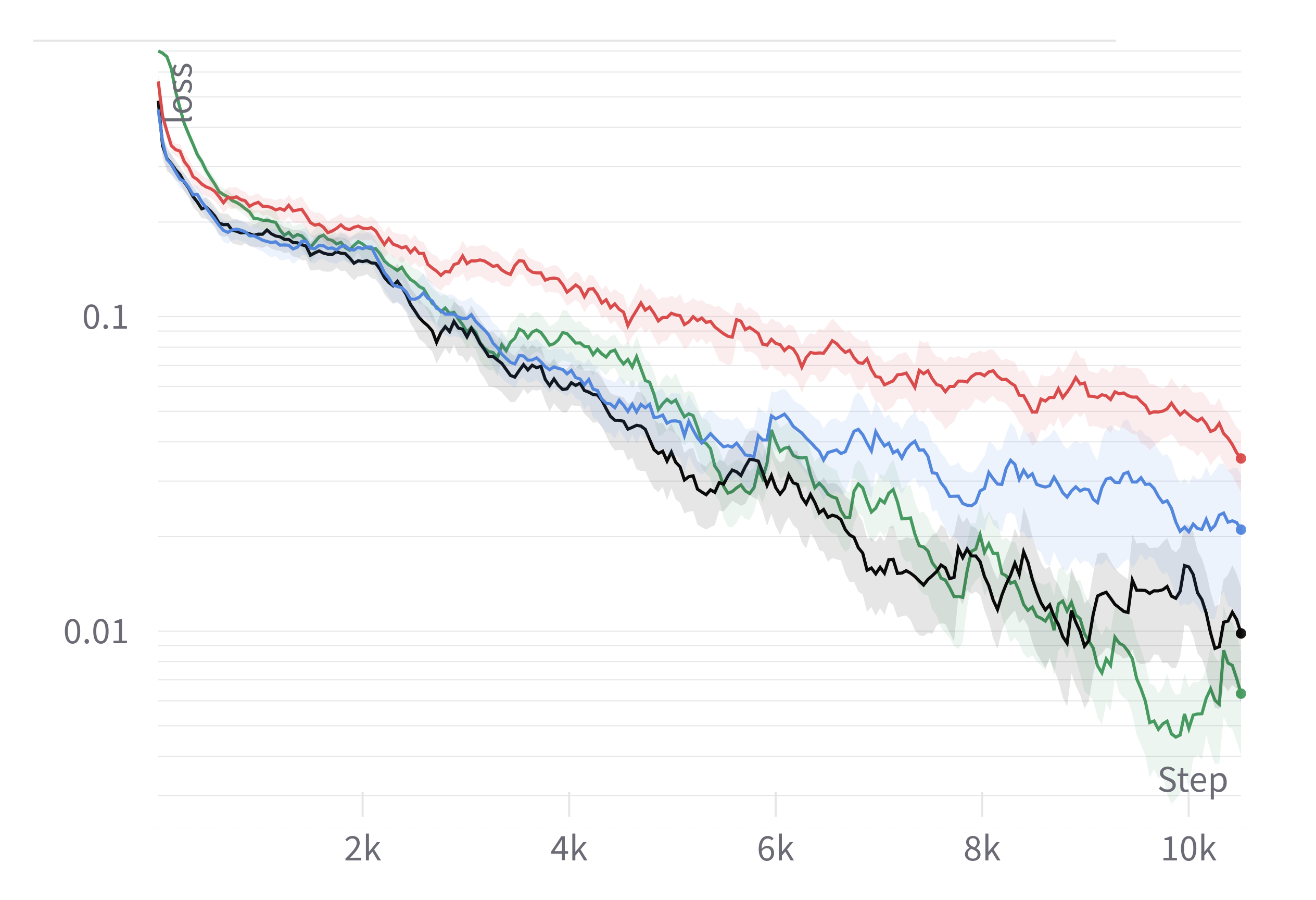}} 
\caption{The loss curves of the experiments on the full  size dataset with standard error indicated around each line. Overall SGDSLS clearly performs worst. In the SST2 experiments PLASLS fails to converge to a very low loss. In the MRPC experiment we can see that ADAMSLS and PLASLS perform better initialy, but ADAM performs about the same in the end.}
\label{fig:lossexp}
\end{figure}

\begin{figure}
\subfloat[MNLI]{\includegraphics[width = 0.25\textwidth]{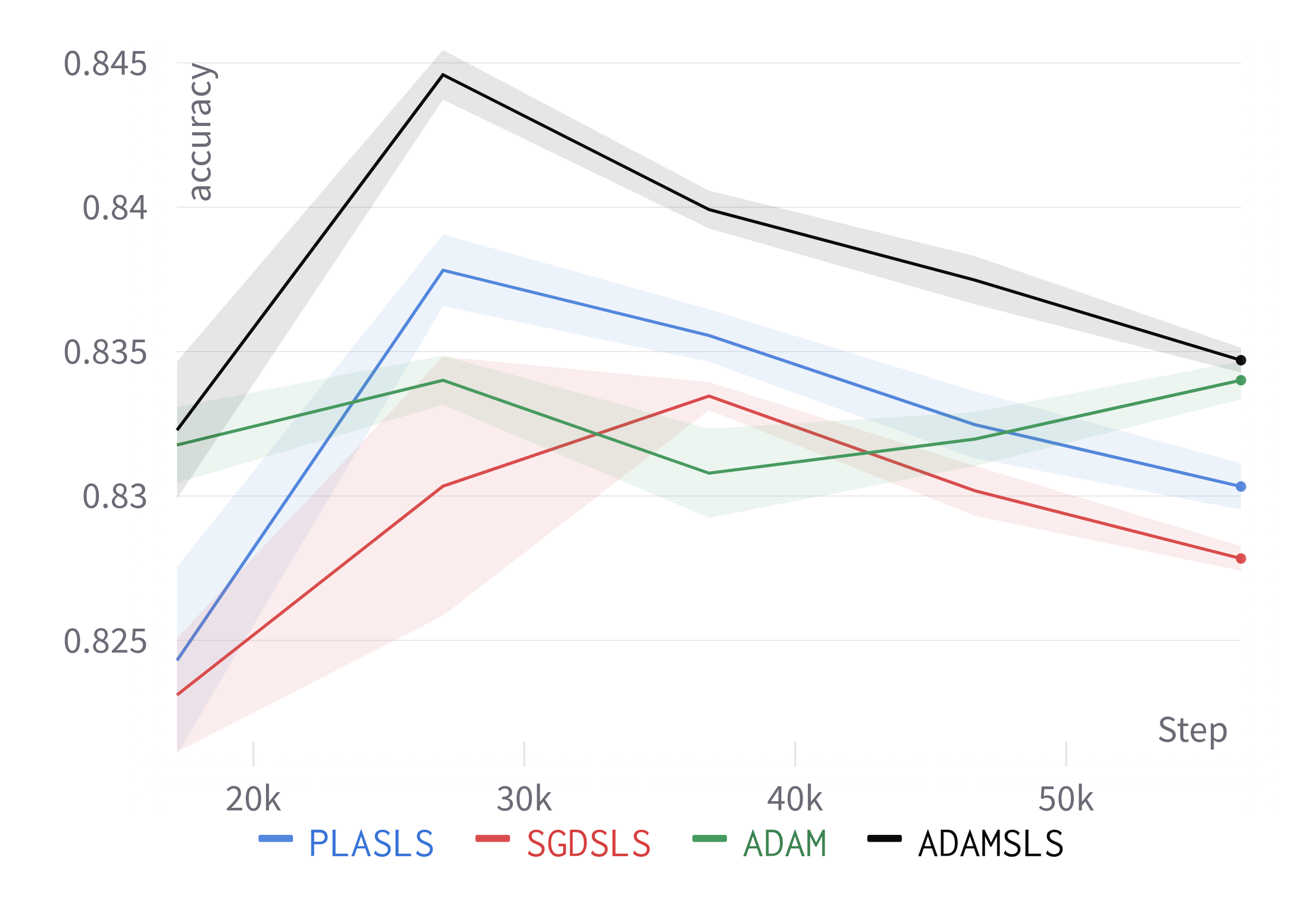}} 
\subfloat[MRPC]{\includegraphics[width = 0.25\textwidth]{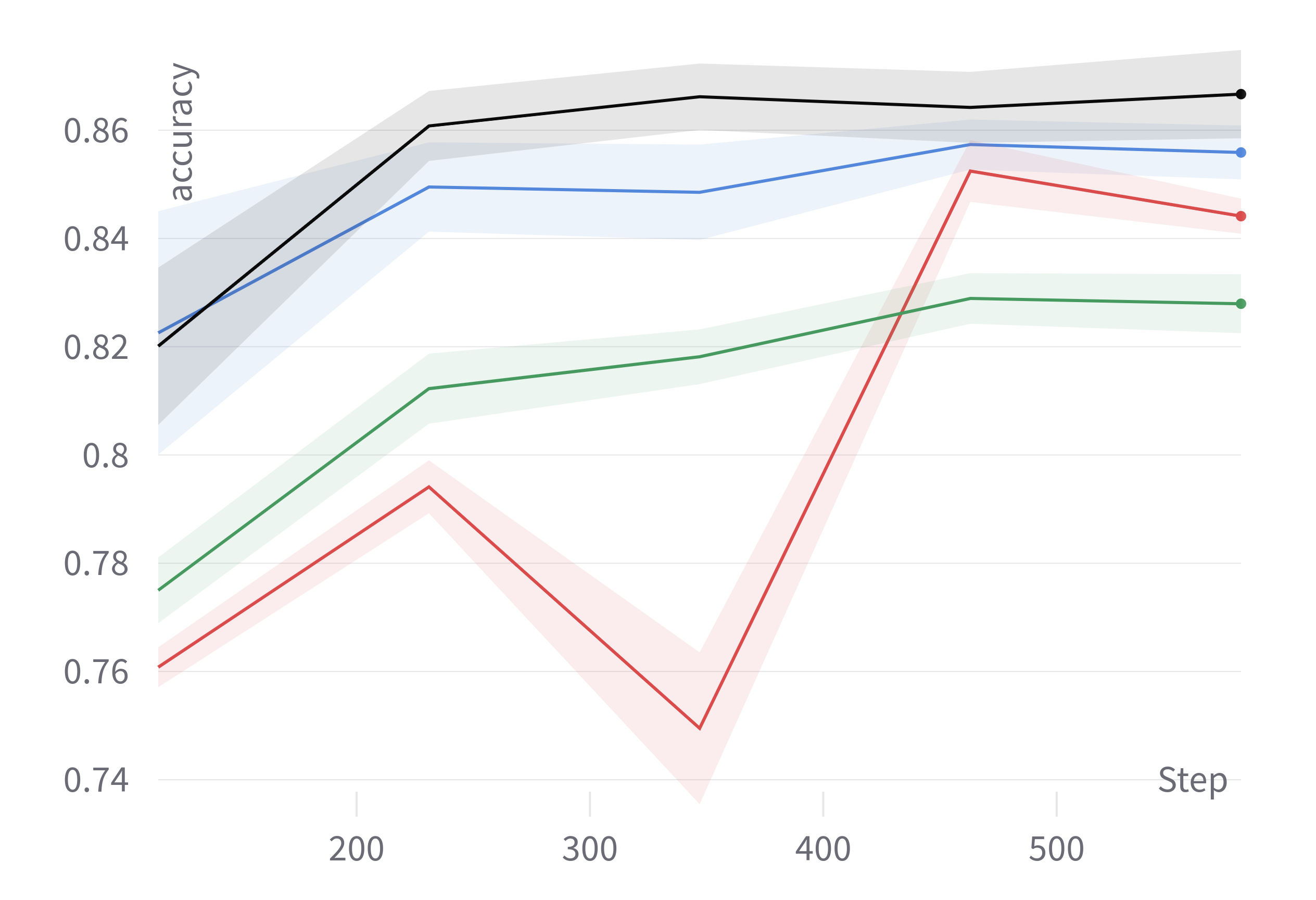}}\\
\subfloat[QNLI]{\includegraphics[width = 0.25\textwidth]{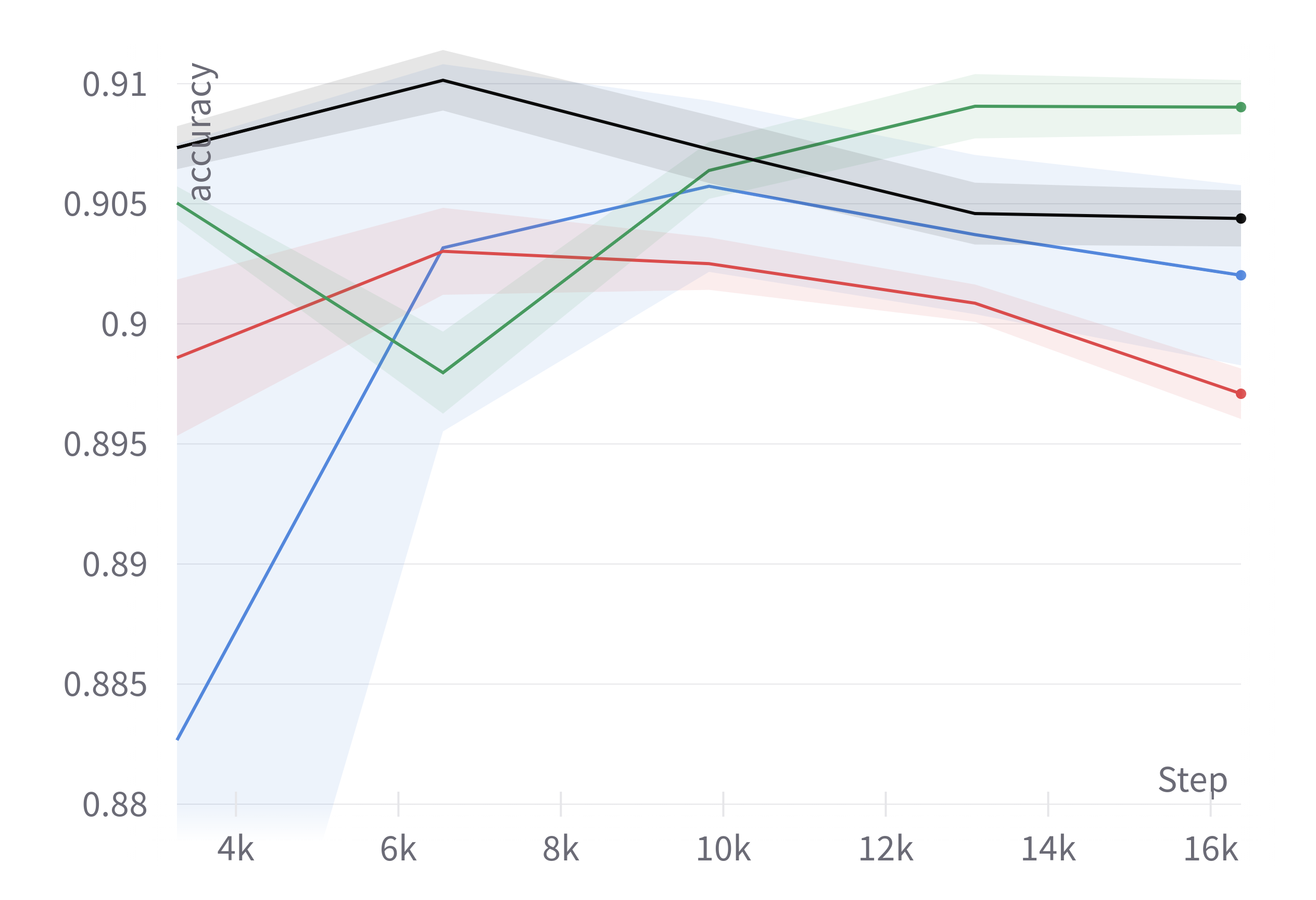}}
\subfloat[SST2]{\includegraphics[width = 0.25\textwidth]{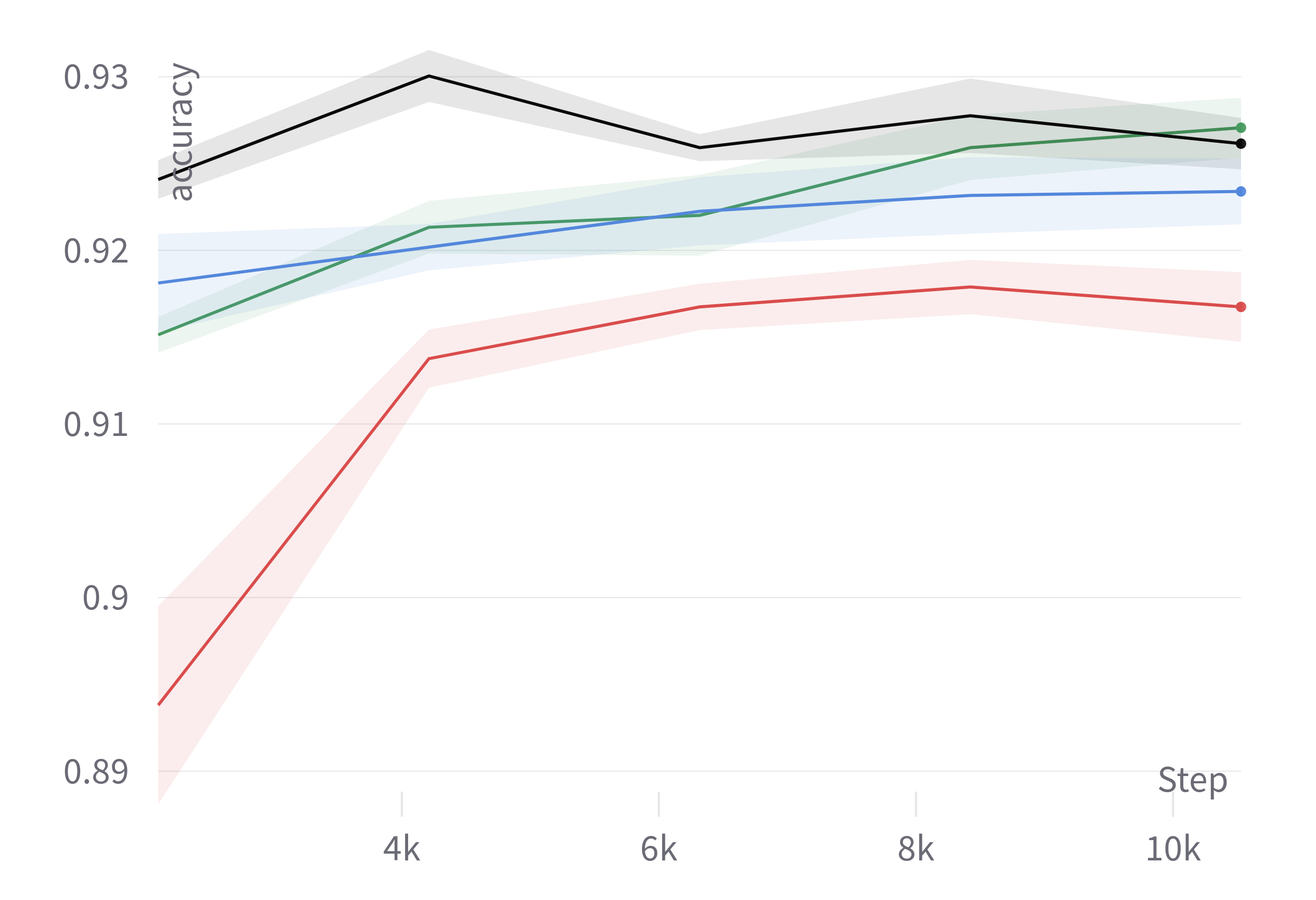}} 
\caption{The accuracy curves of the experiments on the full  size dataset with standard error indicated around each line, starting after the first epoch. Accuracy was calculated on the validation data. Overall we can observe that SGDSLS performs the worst. ADAMSLS is overall the best performing optimizer. }
\label{fig:accexp}
\end{figure}

In this section, we will describe the results of our experiments on the Glue dataset collection. 
We compare the 4 candidates as described in Section \ref{sec:methods}
\begin{itemize}
    \item ADAM
    \item SGDSLS
    \item ADAMSLS
    \item PLASLS
\end{itemize}
All accuracies displayed are the accuracies on the validation sets. The losses displayed are the losses calculated on the training sets, smoothed with exponential moving average. The colored areas around each line indicate the standard error of each experiment. We display the accuracies and losses during the training period in Figures \ref{fig:smallexploss},\ref{fig:smallexpacc},\ref{fig:lossexp},\ref{fig:accexp}. The Tables \ref{Fig:accsmall} and \ref{Fig:acclarge} are displaying the accuracies averaged over all datasets used, as is commonly done for the Glue dataset collection.

\subsection{Small Experiments}

In the small size experiments (see Figure \ref{fig:smallexploss} and Figure \ref{fig:smallexpacc}) we observe that the final loss of ADAMSLS and PLASLS is at least one order of magnitude lower compared to ADAM and SGDSLS. This is also reflected in the accuracy metric where PLASLS and ADAMSLS perform significantly better than ADAM or SGDSLS. Small losses on the training data and high accuracies on the validation data are highly correlated and no overfitting is observed. 

In Table \ref{Fig:accsmall}, we see the average performance over all datasets and runs at the end of training. ADAMSLS and PLASLS clearly outperform ADAM and SGDSLS by about 3\%. Note that this accuracy is taken after 5 epochs, earlier in training even larger advantages for ADAMSLS and PLASLS can be observed.

\begin{table}[b] 
  \centering
  \caption{Average classification accuracies, for the $small$ datasets. Best performing method is marked in \textbf{bold}. }
  \label{Fig:accsmall}
  \begin{tabular}{c cccc}
    \toprule
method & ADAM &  SGDSLS & ADAMSLS & PLASLS \\
 \cmidrule(r){1-1}   \cmidrule(r){2-5} 
$accuracy$ & 0.6875 & 0.6927 & \textbf{0.7250} & 0.7165    \\
    \bottomrule
  \end{tabular}
\end{table}

\subsection{Full Size Experiments}


\begin{table}[b] 
  \centering
  \caption{Average classification accuracies, for the fullsize datasets. Best performing method is marked in \textbf{bold}. }
  \label{Fig:acclarge}
  \begin{tabular}{c cccc}
    \toprule
method & ADAM &  SGDSLS & ADAMSLS & PLASLS \\
 \cmidrule(r){1-1}   \cmidrule(r){2-5} 
$accuracy$ & 0.8745 & 0.8714 & \textbf{0.8830} & 0.8779    \\
    \bottomrule
  \end{tabular}
\end{table}
In the full size experiments, we observe that the loss starts to decrease quicker with ADAMSLS and PLASLS. After about 10k steps ADAM outperforms the other options on the loss metric. On the accuracy metric we can see that ADAMSLS and PLASLS perform better in the beginning, but only slightly better at the end of training.

In Table \ref{Fig:acclarge} we see the final average performance over all datasets and runs. ADAMSLS and PLASLS very slightly outperform ADAM and SGDSLS by about 1\%. It should be noted that these accuracies are measured after 5 epochs. Earlier in training, greater benefits can be observed for ADAMSLS and PLASLS.

\subsection{Discussion}
Overall we see that with smaller training sets or shorter training runs, ADAMSLS and PLASLS perform significantly better than ADAM or SGDSLS. This performance difference vanishes for longer training runs, here ADAM seems to perform better on the loss metric but still performs slightly worse in accuracy.

\begin{figure}
\includegraphics[width = 0.45\textwidth]{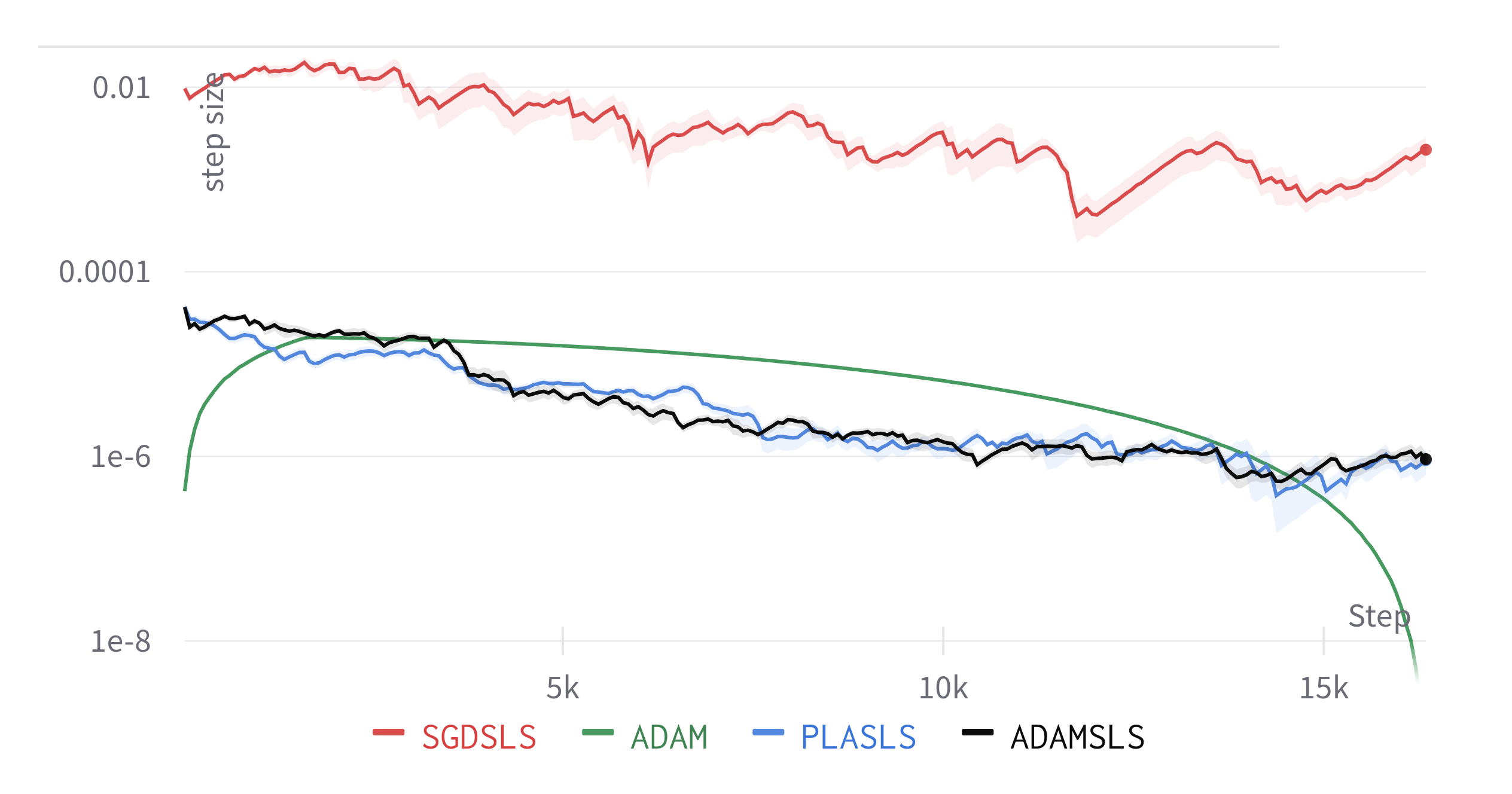}
\caption{Average step size of all layers during training runs on the QNLI dataset.}
\label{fig:avgstep}
\end{figure}

Interestingly, we can see in Figure \ref{fig:avgstep} that the average step size of a well hyperparameter tuned ADAM and the automatically calculated ADAMSLS and PLASLS are similar during long periods of training. This demonstrates, the ability of ADAMSLS and PLASLS to result in very similar step size regimes to the tuned BERT steps size for the Adam optimizer even though they are not given an initial step size and thusly do not need any hyperparameter tuning.

\section{\uppercase{Ablation studies}}
\label{sec:ablation}

In this section we will take a short look at which parts of our PLASLS approach have the most impact on performance, as well as different options for multiple elements of the described algorithm which we tested, but did not go into detail in the other sections.

\subsection{Batch Size}

During our experiments we noticed that the batch size had a large influence on the stability of the training. In this section we compare different batch sizes.

\begin{figure}
\includegraphics[width = 0.45\textwidth]{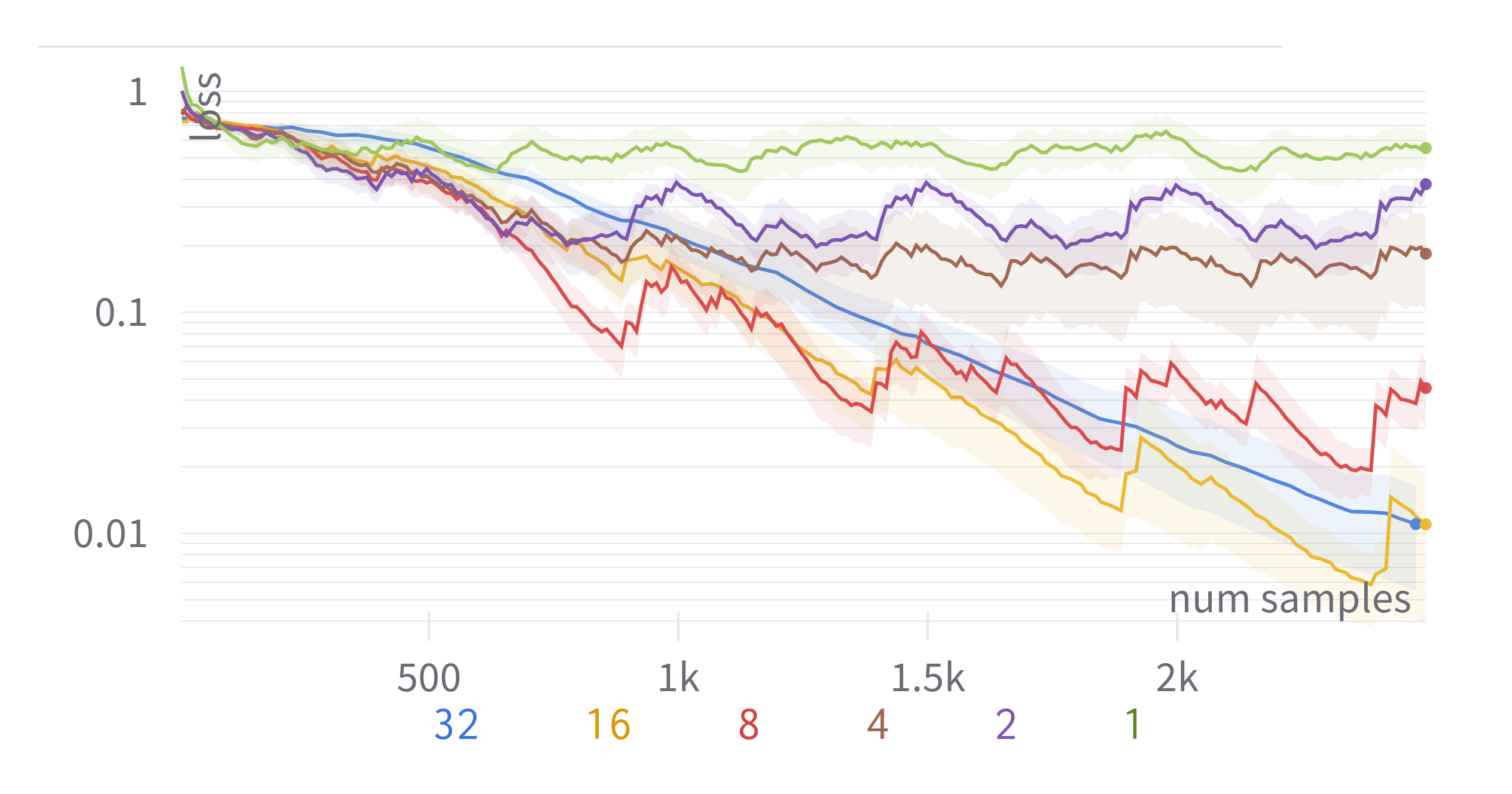}
\caption{Loss on the sst2small dataset for different batch size options for PLASLS. Standard error is visualized around each line. The X-Axis denotes the number of samples processed.}
\label{fig:batchsize}
\end{figure}

In Figure \ref{fig:batchsize}, we see that PLASLS performance improves for larger batch sizes during training. Furthermore, the loss curves get a lot smoother with higher batch sizes. During our experiments we choose a batch size of 32 as it was the highest supported batch size on our available hardware.

\subsection{Split Options}
\begin{figure}
\includegraphics[width = 0.45\textwidth]{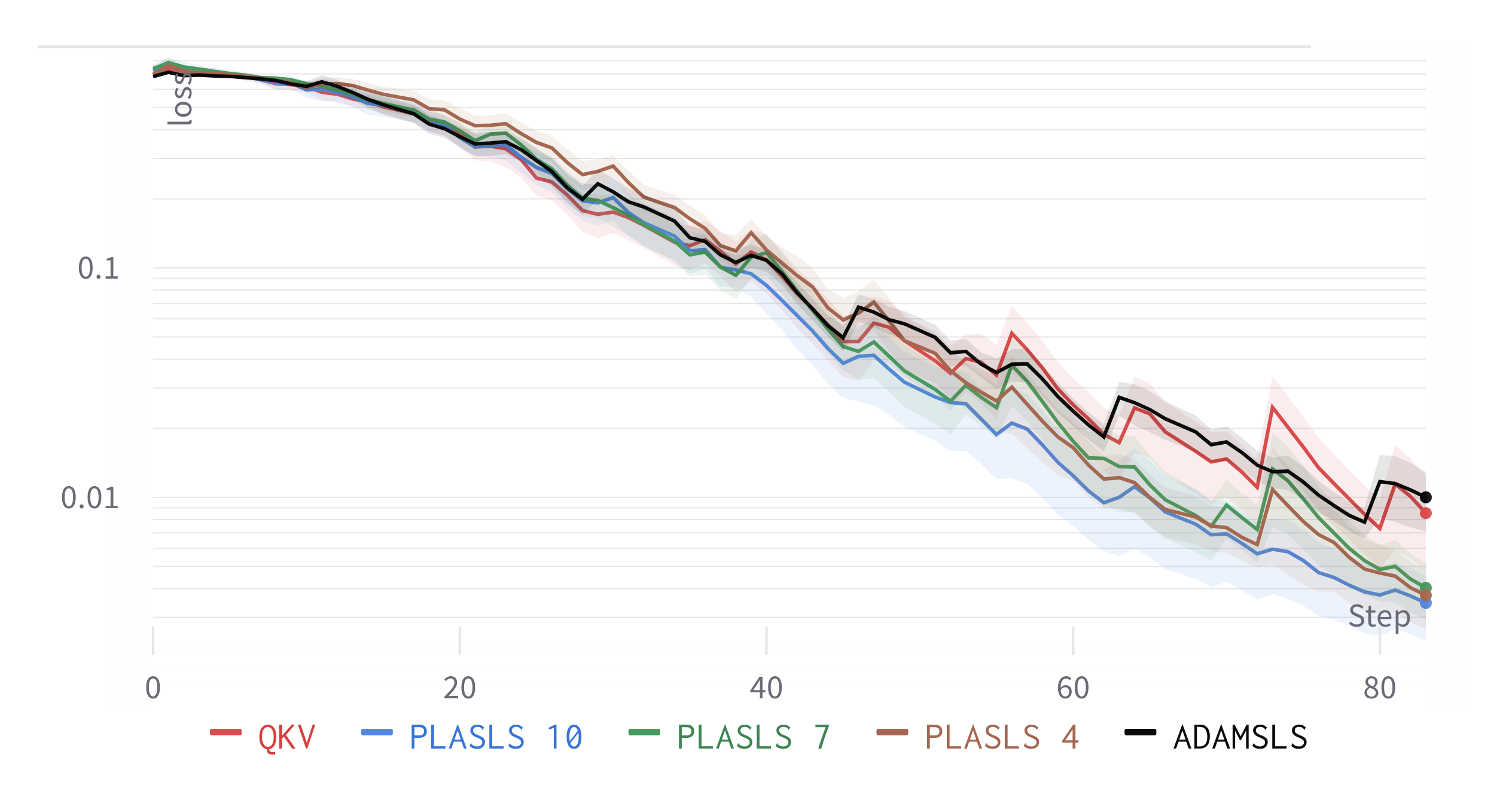}
\caption{Loss on the SST2small dataset for different split options for PLASLS and ADAMSLS as reference. Standard error is visualized around each line.}
\label{fig:layer}
\end{figure}

Several possibilities to split a Transformer based network exist. In this section we look at a split by:
\begin{itemize}
    \item layer 
    \item query, key and value
\end{itemize}
We tried splitting the network in 1,4,7,10 components as well as by query,key and value (QKV). If we split the network in one component, ADAMSLS is equivalent to PLASLS. The loss curves for different layer configurations can be seen in Figure \ref{fig:layer}.
More layers result in gradually better performance for the SST2small dataset.
Splitting by QKV does not seem to improve performance.

\section{\uppercase{Conclusion}}
\label{sec:conclusion}

The type of optimizer, learning rate and learning rate schedule is a critical choice for every machine learning pipeline. We are the first in the literature to evaluate the combination of Adam and line search methods on transformer fine tuning(ADAMSLS) and present a new variant of the line search optimizer PLASLS. 
Both optimizers we presented perform equal or better than state-of-the art baselines, especially on small data sets or short training runs in the domain of natural language processing for the ubiquitous Transformer architecture. We recommend using our ADAMSLS PyTorch implementation as best practice for the task of Transformer fine tuning.

%These optimizers outperform classical state of the art baselines in the domain of natural language processing and Transformers. Especially on smaller datasets and shorter training runs the optimizers presented achieve significant improvements compared to other options such as ADAM or SGDSLS.

%For future work it would be interesting to find a replacement for the Armijo criterion that is less susceptible to the batch size or noise in the training data.

%Another point of possible improvement could be incorporating the constant step sizes of other subnetworks while updating the step size of a single subnetwork, in a changed Armijo criterion.
%TODO further work
For further work it would be interesting to combine the faster initial convergence rate of PLASLS or ADAMSLS with the long-term convergence of ADAM. Either by finding a more stochasticly stable replacement for the Armijo criterion, or by manually switching to the Adam optimizer after a certain amount of steps. By combining/improving the optimizer thusly, we could get an optimizer which would be even more widely applicable.
%\noindent

The source code is open-source and free (MIT licensed) software and available at \newline \url{https://github.com/TheMody/Faster-Convergence-for-Transformer-Fine-tuning-with-Line-Search-Methods}.

\bibliographystyle{ieeetr}
\bibliography{references}

\end{document}